\documentclass[10pt,logo,copyright]{nvidia}

\usepackage{color,xcolor}
\usepackage{epsfig}
\usepackage{graphicx}

\usepackage{adjustbox}
\usepackage{array}
\usepackage{booktabs}
\usepackage{colortbl}
\usepackage{float,wrapfig}
\usepackage{float}
\usepackage{hhline}
\usepackage{multirow}
\usepackage{subcaption} %
\usepackage[toc,page]{appendix}
\usepackage{stfloats}
\usepackage{enumitem}
\usepackage{paralist}

\usepackage{amsmath,amsfonts,amsthm,amssymb}
\usepackage{bm}
\usepackage{nicefrac}
\usepackage{microtype}
\usepackage{inconsolata}
\usepackage{pifont}
\usepackage{bbm}

\usepackage{changepage}
\usepackage{extramarks}
\usepackage{fancyhdr}
\usepackage{lastpage}
\usepackage{setspace}
\usepackage{soul}
\usepackage{xspace}
\usepackage{indentfirst}
\usepackage{paralist}
\usepackage{booktabs,arydshln}
\usepackage{makecell}

\definecolor{iccvblue}{rgb}{0.21,0.49,0.74}
\usepackage[pagebackref,breaklinks,colorlinks,allcolors=iccvblue]{hyperref}
\usepackage{url}

\usepackage{algorithm, algorithmic}
\usepackage{enumerate}
\usepackage{lipsum}
\usepackage{minted}
\usepackage{tikz}
\usetikzlibrary{tikzmark}
\usepackage{xfrac}

\newcolumntype{L}[1]{>{\raggedright\let\newline\\\arraybackslash\hspace{0pt}}m{#1}}
\newcolumntype{C}[1]{>{\centering\let\newline\\\arraybackslash\hspace{0pt}}m{#1}}
\newcolumntype{R}[1]{>{\raggedleft\let\newline\\\arraybackslash\hspace{0pt}}m{#1}}

\newcommand{\sect}[1]{Section~\ref{#1}}

\newcommand{\fig}[1]{Figure~\ref{#1}}

\newcommand{\tab}[1]{Table~\ref{#1}}

\newcommand{\ignorethis}[1]{}

\makeatletter
\DeclareRobustCommand\onedot{\futurelet\@let@token\@onedot}
\def\@onedot{\ifx\@let@token.\else.\null\fi\xspace}

\def\eg{\emph{e.g}\onedot} 
\def\ie{\emph{i.e}\onedot}

\makeatother

\makeatletter
\def\adl@drawiv#1#2#3{%
        \hskip.5\tabcolsep
        \xleaders#3{#2.5\@tempdimb #1{1}#2.5\@tempdimb}%
                #2\z@ plus1fil minus1fil\relax
        \hskip.5\tabcolsep}
\newcommand{\cdashlinelr}[1]{%
  \noalign{\vskip\aboverulesep
           \global\let\@dashdrawstore\adl@draw
           \global\let\adl@draw\adl@drawiv}
  \cdashline{#1}
  \noalign{\global\let\adl@draw\@dashdrawstore
           \vskip\belowrulesep}}
\makeatother

\definecolor{citecolor}{HTML}{0071bc}
\definecolor{mydarkblue}{rgb}{0,0.08,1}
\definecolor{mydarkgreen}{rgb}{0.02,0.6,0.02}
\definecolor{mydarkred}{rgb}{0.8,0.02,0.02}
\definecolor{mydarkorange}{rgb}{0.40,0.2,0.02}
\definecolor{mypurple}{RGB}{111,0,255}
\definecolor{myred}{rgb}{1.0,0.0,0.0}
\definecolor{mygold}{rgb}{0.75,0.6,0.12}
\definecolor{mydarkgray}{rgb}{0.66, 0.66, 0.66}

\definecolor{darkblue}{rgb}{0,0.08,1}
\definecolor{darkgreen}{rgb}{0.02,0.6,0.02}
\definecolor{darkred}{rgb}{0.8,0.02,0.02}
\definecolor{darkorange}{rgb}{0.40,0.2,0.02}
\definecolor{darkpurple}{RGB}{111,0,255}

\def\method{SparseVILA\xspace}

\begin{document}

\title{\method: Decoupling Visual Sparsity for Efficient VLM Inference}

\author{Samir Khaki\textsuperscript{4} \qquad Junxian Guo\textsuperscript{2} \qquad Jiaming Tang\textsuperscript{2} \qquad Shang Yang\textsuperscript{2} \qquad Yukang Chen\textsuperscript{1} \qquad Konstantinos N. Plataniotis\textsuperscript{4} \qquad Yao Lu\textsuperscript{1} \qquad Song Han\textsuperscript{1,2} \qquad Zhijian Liu\textsuperscript{1,3} \\~\\
\textsuperscript{1}NVIDIA \quad \textsuperscript{2}MIT \quad 
\textsuperscript{3}UC San Diego \quad 
\textsuperscript{4}University of Toronto}

\begin{abstract}

\textbf{Abstract:} \hspace{2pt} Vision Language Models (VLMs) have rapidly advanced in integrating visual and textual reasoning, powering applications across high-resolution image understanding, long-video analysis, and multi-turn conversation. However, their scalability remains limited by the growing number of visual tokens that dominate inference latency. We present \textbf{\method}, a new paradigm for efficient VLM inference that \textit{decouples} visual sparsity across the prefilling and decoding stages. SparseVILA distributes sparsity across stages by pruning redundant visual tokens during prefill and retrieving only query-relevant tokens during decoding. This decoupled design matches leading prefill pruning methods while preserving multi-turn fidelity by retaining most of the visual cache so that query-aware tokens can be retrieved at each conversation round. Built on an AWQ-optimized inference pipeline, SparseVILA achieves up to \textbf{4.0$\times$ faster prefilling}, \textbf{2.5$\times$ faster decoding}, and an overall \textbf{2.6$\times$ end-to-end speedup} on long-context video tasks -- while improving accuracy on document-understanding and reasoning tasks. By decoupling query-agnostic pruning and query-aware retrieval, SparseVILA establishes a new direction for efficient multimodal inference, offering a training-free, architecture-agnostic framework for accelerating large VLMs without sacrificing capability.

\vspace{10pt}

\begin{center}
    \includegraphics[width=\linewidth]{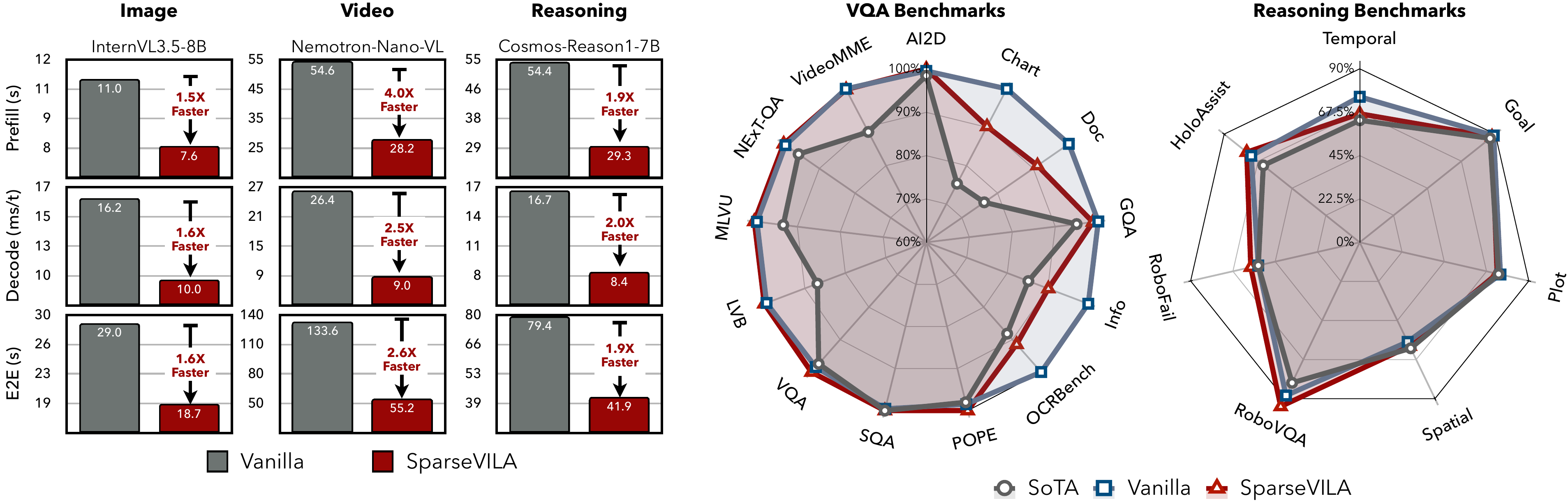}
    \captionof{figure}{\textbf{Figure 1 | SparseVILA – Efficient VLM Inference.}  (a) SparseVILA delivers consistent speedups across image, video, and reasoning tasks, with up to \textbf{4.0$\times$} gains in the prefill stage, \textbf{2.5$\times$} gains in decoding throughput, and \textbf{2.6$\times$} end-to-end speedup. (b) SparseVILA maintains competitive accuracy across most VQA and reasoning benchmarks. While document understanding tasks show a modest drop, performance remains above \textbf{90\%} of the best reported scores, whereas other compression methods often fall below \textbf{75\%}. Inference speed in (a) is measured using a single NVIDIA A6000 GPU. Accuracy values in (b) are normalized relative to the highest score for each benchmark.}
    \label{fig:teaser}
\end{center}

\end{abstract}

\maketitle

\section{Introduction}

Vision Language Models (VLMs) have emerged as a state-of-the-art conversational tool enabling users to directly interact with Large Language Models (LLMs) using various visual features, including photographs, documents, and videos~\cite{liu2024llava, lin2024vila, bai2023qwenvl, wang2024qwen2vl, liu2025nvila}. Unfortunately, this added modality comes at the expense of higher latency and memory associated with processing the visual tokens in the LLM. Hence, deploying LLMs efficiently at inference time remains a challenge.

Several works have aimed to reduce these associated costs through model pruning on the LLM or vision encoder~\cite{sun2024wanda, ma2023llmpruner, zhu2021vtp}, KV cache compression~\cite{lin2025qserve, hooper2024kvquant, liu2025kivi}, and most recently, token sparsification~\cite{chen2024fastv, yang2025visionzip}. By reducing the amount of computation in the inference pipeline, many of these methods can achieve significant context stage savings, as this stage of the network is mainly compute-bound. %

Looking beyond the context, real-world applications often demand extensive generation. Tasks such as image captioning may require a few hundred tokens; meanwhile, video captioning or detailing easily requires more than a few thousand generated tokens. Hence, in such applications, it is not sufficient to only focus on context stage optimization -- efficient implementations with real-world applications should focus on both context and decoding stage optimizations. One such example is a multi-turn conversation.

Multi-turn/round conversation serves as a practical use case for VLMs, wherein a user may pose multiple questions about a given visual input. In fact, most benchmarks inherently support multi-round conversation: the GQA dataset~\cite{hudson2019gqa} has more than 90 questions for the same visual input. Despite this, most evaluation benchmarks run single-round evaluation (i.e., repetitive pre-filling), which is not only unrealistic but also inefficient, as the context stage would be repeated for each generation round. In real-world scenarios, the visual input could span tens of thousands of context tokens; hence, repetitive pre-filling would dramatically slow down the user's interaction with their VLM. 

In this work, we aim to present a unified approach for tackling context and decoding latency in modern VLMs. Latency is a common challenge associated with VLMs due to the sheer amount of visual tokens to be processed. Existing methods for accelerating VLM inference primarily focus on token-wise pruning or merging techniques, with recent approaches leveraging textual priors to reduce visual token complexity in a query-aware manner. In practice, methods that permanently remove visual tokens during the context stage are quite lossy in multi-turn evaluations, as visualized in \fig{fig:teaser}. Hence, in this paper, we introduce \method as a novel approach for accelerating VLM inference, while retaining multi-turn performance. 

Our key insight is a \textit{decoupled sparsity framework} enabling \method to migrate sparsity from the prefill into the decoding stage. Further, \method leverages query-aware retrieval in the decoding stage, supporting multi-turn conversation as a different subset of context tokens can be retrieved per question. This \textit{decoupled} approach allows \method to achieve significant performance improvements in image-centric benchmarks, as shown in \fig{fig:teaser} and outperforms previous methods in long-context/generation scaling.

\section{Preliminaries}
\label{sect:prelim}

Token pruning has proven effective in accelerating inference across a variety of tasks, including image classification~\cite{bolya2023tome, kong2022spvit, khaki2024optin}, object detection~\cite{chen2023sparsevit}, and semantic segmentation~\cite{chen2023sparsevit, tang2023dtop}. With the rise of generative AI, these techniques have been further extended to diffusion models~\cite{bolya2023tomesd}, large language models~\cite{kim2022ltp}, and vision-language models~\cite{zhang2025sparsevlm, yang2025visionzip, huang2024ivtp}. We refer the readers to \sect{sect:related} for a detailed survey of related work.

This paper focuses on token pruning/sparsity for vision-language models (VLMs). The key idea is that \textit{not all visual tokens contribute equally} to VLM's final prediction. By identifying and removing less informative tokens, it is possible to significantly reduce the computational cost of VLM inference and thereby improve efficiency. Existing work in this area largely differs in how these tokens are selected. We categorize prior methods into two groups based on their dependence on the input query: (i) \textit{query-agnostic} approaches, which identify unimportant tokens based solely on visual saliency or redundancy, and (ii) \textit{query-aware} approaches, which incorporate the semantic relationship between visual and textual inputs to guide pruning. For each category, we highlight representative methods and their limitations, laying the groundwork for our method.

\subsection{Query-Agnostic Sparsity}
\label{sec:background:query-agnostic}

Query-agnostic token pruning methods aim to reduce redundancy or select important visual tokens without relying on the textual input (\ie, query). They prune tokens based solely on the visual context, either within or after the vision encoder. For example, PruMerge~\cite{shang2025prumerge} clusters and discards less informative tokens using final-layer attention scores, while VisionZip~\cite{yang2025visionzip} employs a token merging module, similar to ToMe~\cite{bolya2023tome}, to compress redundant visual information.

This visual-only focus, however, presents key limitations. First, these methods often sacrifice fine-grained visual details, especially under high sparsity, which can degrade performance. More importantly, they cannot adapt token selection based on the input query, leading to suboptimal results when task-relevant information is sparsely distributed. 
By treating all visual tokens uniformly, they risk discarding critical information necessary for accurate reasoning.

\subsection{Query-Aware Sparsity}
\label{sec:background:query-aware}

Query-aware pruning improves visual token selection by explicitly modeling the relationship between textual queries and visual representations. For example, FastV~\cite{chen2024fastv} leverages attention maps from early LLM layers as salience indicators to guide token pruning during the prefill stage, while SparseVLM~\cite{zhang2025sparsevlm} uses query-to-vision attention to discard less relevant visual tokens.

Although effective for single-turn tasks, query-aware pruning faces notable limitations in multi-turn interactions. Pruning decisions made for an initial query can permanently remove visual information crucial for subsequent questions, leading to degraded performance across conversation rounds. Empirically, such methods show sharp accuracy drops in multi-turn dialogue, often underperforming even query-agnostic baselines.

\begin{figure}[t]
    \centering
    \begin{minipage}[t]{0.62\linewidth}
        \centering
        \includegraphics[width=\linewidth]{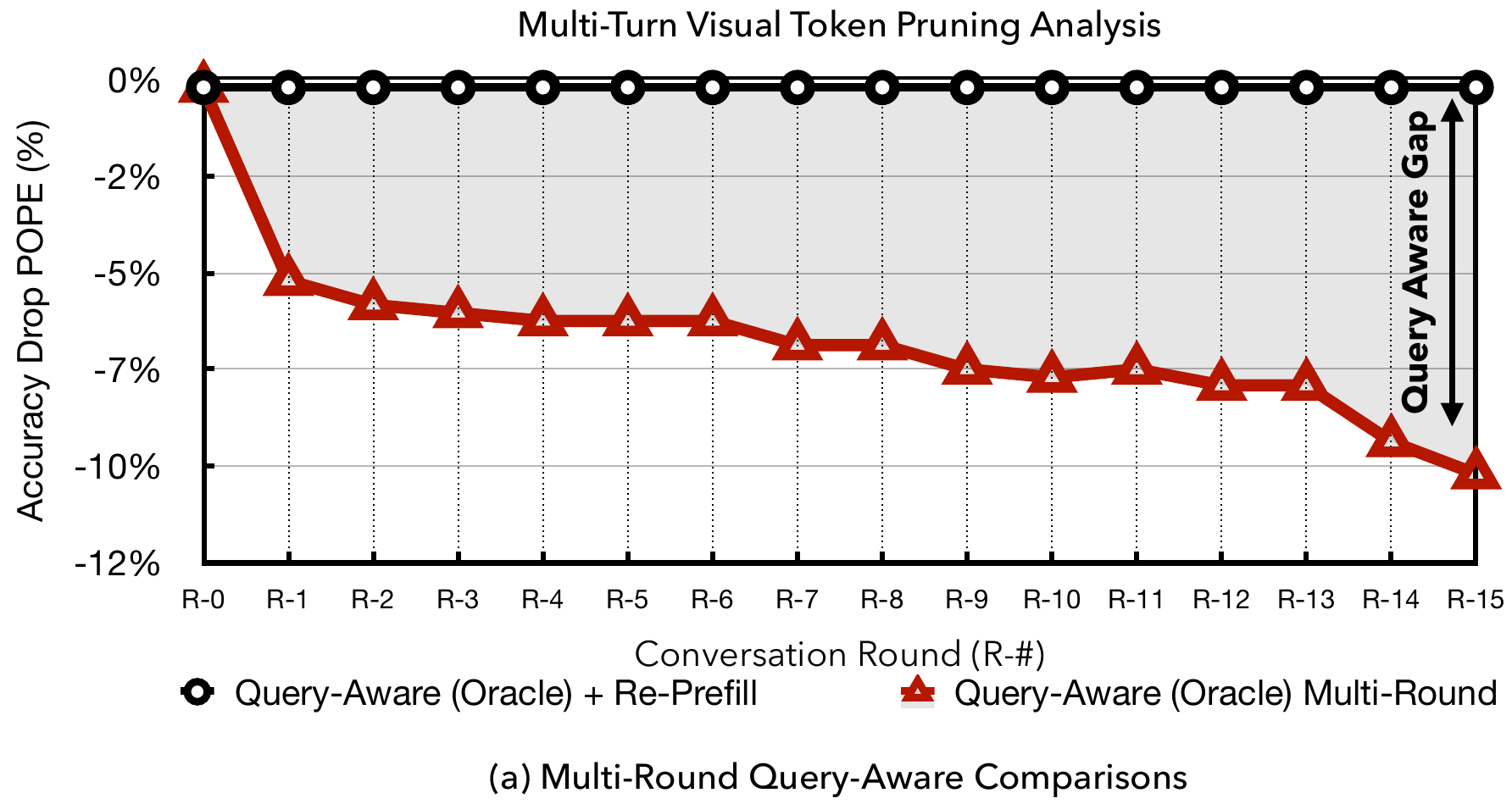}
        \caption{Multi-Round Query-Aware Comparisons with LLaVA-1.5~\cite{liu2024llava} on the POPE~\cite{li2023pope} dataset. Without re-prefilling the context, the query-aware oracle degrades heavily, indicating the inability of query-aware pruning to scale effectively in a multi-turn conversation.}
        \label{fig:query_oracle_degradation}
    \end{minipage}%
    \hfill
    \begin{minipage}[t]{0.35\linewidth}
        \centering
        \includegraphics[width=\linewidth]{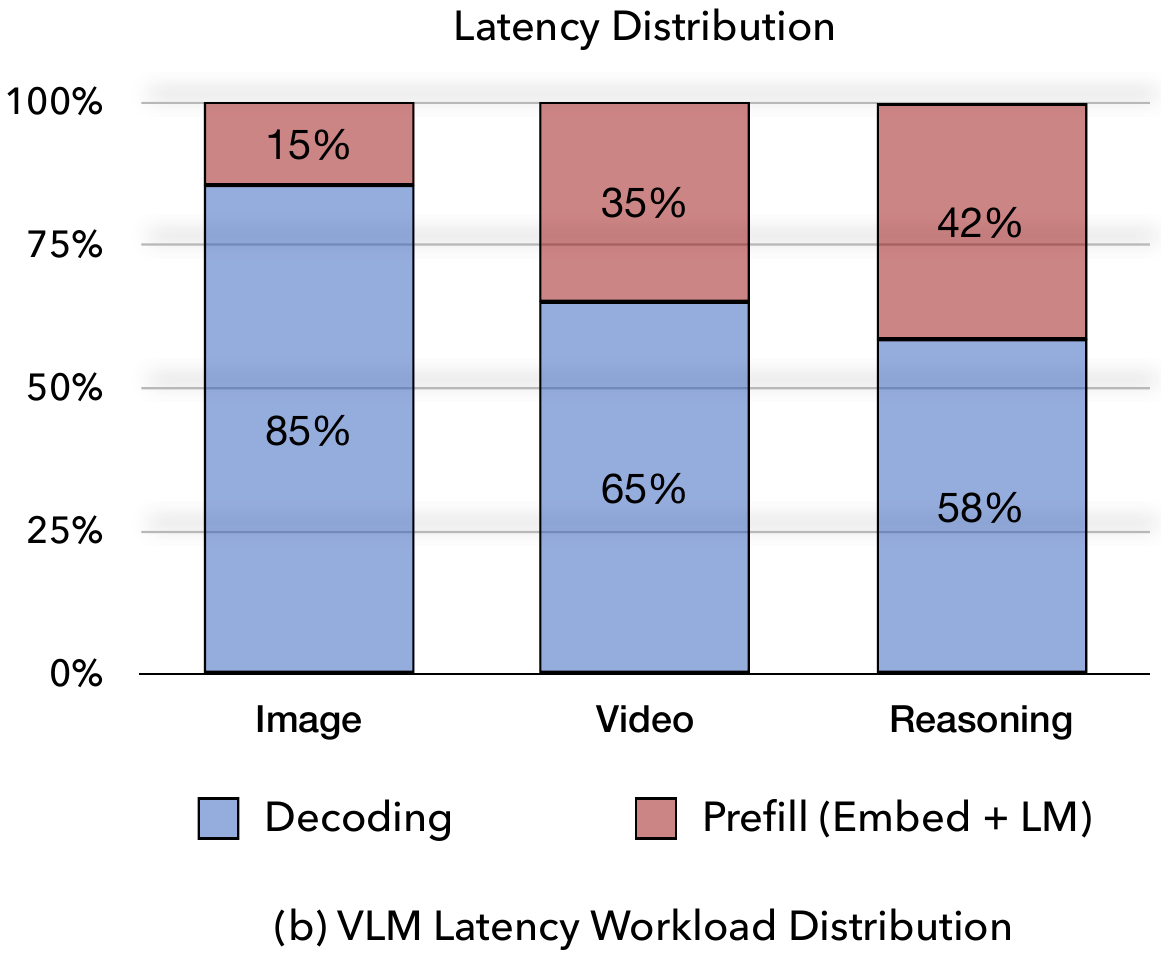}
        \caption{Latency distribution over the prefilling and decoding stages across image, video, and reasoning workloads.}
        \label{fig:latency_distribution}
    \end{minipage}
\end{figure}

To examine this limitation, we construct a \textit{query-aware oracle} that greedily selects an optimal subset of visual tokens to maximize agreement with the unpruned model's responses. The oracle represents the theoretical upper bound for any query-aware approach, as it directly leverages both the current query and the ground-truth response during selection. Yet, as shown in \fig{fig:query_oracle_degradation}, even this oracle exhibits substantial degradation over successive conversation rounds, highlighting a fundamental constraint of query-dependent pruning: once informative tokens are removed, they cannot be recovered in later turns. These findings motivate the need for a decoupled sparsity framework that preserves visual coverage during prefill while allowing query-aware retrieval during decoding.

\section{\method: Best of Both Worlds}

\begin{figure*}[t]
    \centering
    \includegraphics[width=\linewidth]{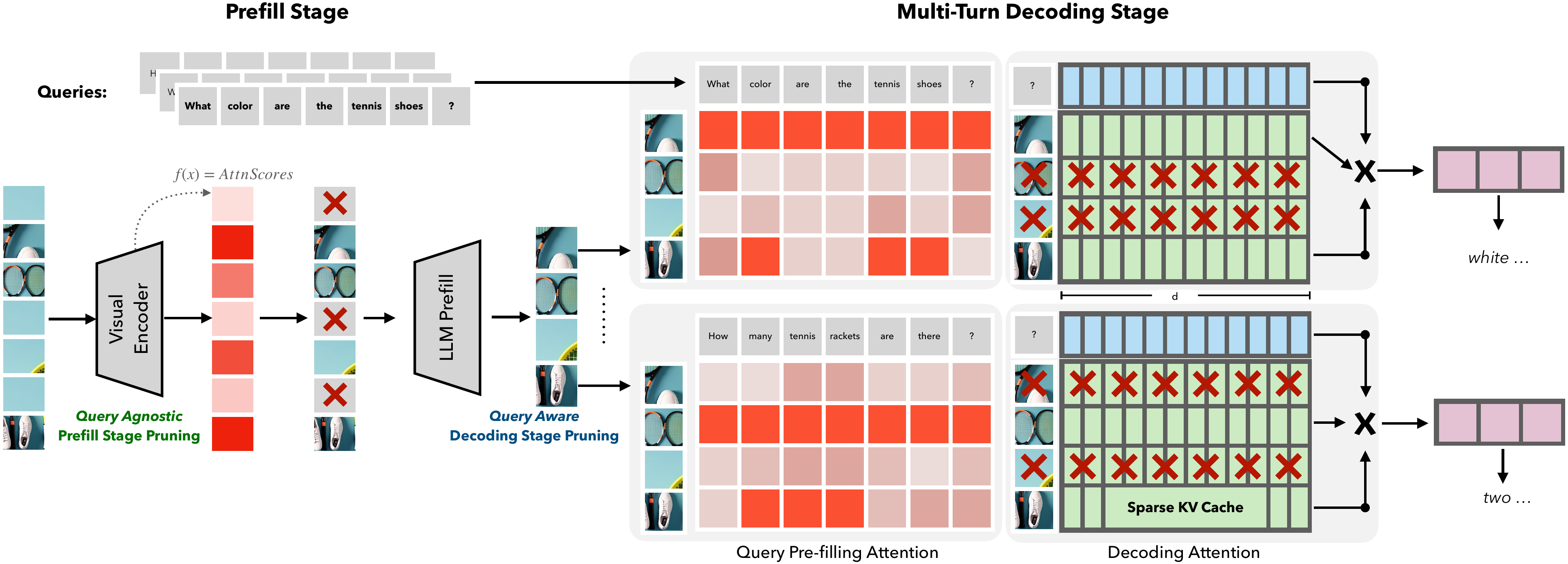}
    \caption{Overview of SparseVILA’s decoupled sparsity framework. In the prefill stage, query-agnostic pruning removes redundant visual tokens based on salience scores from the visual encoder, yielding a compact representation shared across conversation turns. During decoding, query-aware retrieval selects only the most relevant visual tokens from the KV cache for attention computation, accelerating generation while maintaining multi-turn fidelity.}
    \label{fig:method}
\end{figure*}

Query-agnostic and query-aware pruning offer complementary benefits but also have inherent limitations. Query-agnostic methods efficiently remove redundant visual tokens without requiring text input, making them stable across multi-turn dialogue, yet they fail to adapt to query-specific relevance. In contrast, query-aware methods dynamically align visual attention with the current query, improving single-turn reasoning, but suffer from irreversible information loss and degraded performance in later conversation rounds once tokens are pruned. These opposing trade-offs motivate our approach. The key insight is that visual sparsity should not be applied uniformly across the inference pipeline. Instead, it should adapt to the distinct roles of the \textit{prefill} and \textit{decoding} stages: the former constructs the multimodal context once, while the latter dominates overall latency during iterative generation.

In this paper, we introduce \method, a framework that achieves the \textit{best of both worlds} by \textbf{decoupling} visual compression across the two stages. \method performs lightweight, query-agnostic pruning during prefill to reduce redundancy without sacrificing coverage, and applies aggressive, query-aware retrieval during decoding when the question is known. This design also better matches the decoding-heavy latency profile of modern VLMs (see \fig{fig:latency_distribution}), yielding significant speedups while maintaining high accuracy across image, video, and reasoning tasks. By separating \textit{when} and \textit{how} sparsity is applied, \method preserves contextual grounding for future turns and enables efficient, query-conditioned reasoning.

\begin{figure}
    \centering
    \includegraphics[width=\linewidth]{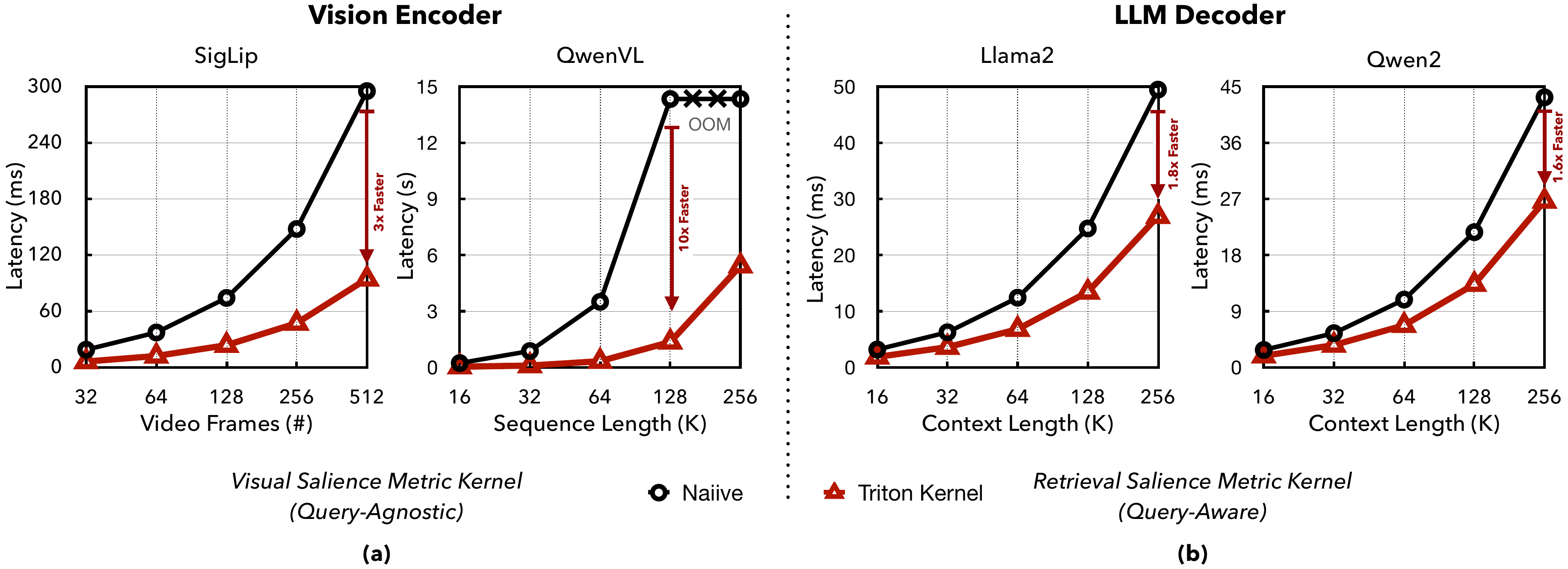}
    \caption{\textbf{Salience Metric Kernels.} Latency comparison between the naïve and custom Triton implementations across two settings: \textbf{(a)} query-agnostic salience computation for the SigLIP and QwenVL vision encoders, and \textbf{(b)} query-aware retrieval salience for the Llama2 and Qwen2 decoder backbones. Our custom kernels consistently accelerate both query-agnostic and retrieval salience computations, achieving up to 10$\times$ and 1.8$\times$ speedups, respectively.}
    \label{fig:triton_kernel}
\end{figure}

\subsection{Prefill Phase: Query-Agnostic Pruning}
\label{sec:method:context}

During the prefill stage, the vision–language model (VLM) encodes the system prompt, visual tokens, and optionally the first user query to construct the multimodal context. To ensure stable performance across multiple dialogue turns, pruning at this stage must remain strictly \textit{query-agnostic}—guided only by visual redundancy or salience rather than any text-conditioned correlation. Since the visual context is computed once and reused throughout the conversation, pruning must retain sufficient coverage for future queries while minimizing redundant information.

\paragraph{Token Salience Estimation.}
We estimate token importance directly from the visual encoder’s self-attention maps, providing a query-independent measure of visual salience. Following prior work~\cite{yang2025visionzip, shang2025prumerge, arif2025hired}, we aggregate attention signals to quantify each token’s contribution to the overall representation, pruning those with the lowest aggregate salience. For models with a single summary token (e.g., CLIP), salience is defined by each token’s attention contribution to this global embedding. For encoders such as RADIO~\cite{ranzinger2024amradio, heinrich2025radio25}, which employ multiple summary tokens, we compute salience as the mean attention directed toward these summary tokens, effectively capturing the same global aggregation behavior. For models without summary tokens (e.g., SigLIP, QwenVL), importance is estimated by averaging intra-visual attention across all tokens.

\paragraph{Efficient Implementation.}
For long-context inputs such as video sequences, attention-based salience estimation can be memory- and latency-intensive. To address this, we implement a custom Triton~\cite{tillet2019triton} kernel that streams softmax normalization and salience accumulation without explicitly forming the full attention matrix. This enables efficient salience computation even for hundreds of thousands of tokens. Empirically, the kernel yields up to a \textbf{3$\times$} acceleration for SigLIP-style encoders and up to \textbf{10$\times$} for QwenVL-style encoders (\fig{fig:triton_kernel}a), forming the computational foundation for \method's scalable prefill pruning.

\subsection{Decode Phase: Query-Aware Retrieval}
\label{sec:method:decoding}

During the decoding phase, the VLM becomes memory-bound as it repeatedly computes next-token predictions using the pre-filled KV cache. To accelerate this process, \method selectively activates only the most query-relevant visual tokens during decoding attention, while preserving the rest of the visual information in the KV cache for potential use in later turns. This design enables query-conditioned sparsity without permanently discarding context, maintaining the flexibility required for multi-turn reasoning.

\paragraph{Query-Aware Token Selection.}
Before decoding begins, \method estimates the relevance of each visual token to the current query using attention-based salience. Specifically, it measures the aggregate attention strength between the query embeddings and visual entries in the KV cache, providing a query-aware signal that highlights which tokens the model is most likely to reference during generation. Tokens with the highest relevance scores are retained for decoding, while less relevant tokens remain cached but inactive. This dynamic retrieval process effectively narrows the attention scope to the most informative subset of visual tokens, improving efficiency without compromising context consistency. We extend the Triton kernel from the prefill stage to stream the relevance computation directly between the query and cached visual tokens. This operation executes concurrently with the FlashAttention2~\cite{dao2024flashattention2} path during prefill, yielding up to a \textbf{1.5$\times$} speedup over a naïve implementation (\fig{fig:triton_kernel}b). Once salience scores are obtained, the selected visual KV entries are compactly packed into a contiguous memory region, avoiding irregular sparse access patterns during autoregressive decoding.

\paragraph{Rotary Embeddings.}
Modern VLMs employ rotary position embeddings (RoPE) to encode positional information across modalities. Conventional architectures such as LLaVA-NeXT and LongVILA apply a unified RoPE to both text and vision tokens, while newer models like Qwen2.5-VL use multimodal RoPE to maintain distinct positional grids. When pruning tokens across prefill and decoding stages, these positional structures can become misaligned. For models using unified RoPE, we simply retain a contiguous range of position indices corresponding to the preserved visual tokens. For multimodal RoPE, we reconstruct the minimal contiguous positional grid along temporal, height, and width dimensions and then shift subsequent text positions to maintain global continuity. This adjustment ensures consistent cross-modal alignment even under aggressive token compression, preserving the integrity of the shared positional embedding space across the entire KV cache.

\subsection{Decoupled Prefill–Decode Visual Sparsity}

\method introduces a decoupled sparsity framework that explicitly separates \textit{where} and \textit{how} visual compression is applied across the inference pipeline. This design is motivated by the distinct computational characteristics of the two stages: the \textit{prefill stage} executes once per visual input to build the multimodal context, while the \textit{decoding stage} performs iterative next-token prediction and typically dominates end-to-end latency (Fig.~\ref{fig:query_oracle_degradation}). Applying uniform sparsity across both is therefore suboptimal—aggressive prefill pruning can permanently discard visual information required for later turns, whereas decoding remains the primary runtime bottleneck.

To address this imbalance, \method decouples sparsity between stages: lightweight, query-agnostic pruning is applied during prefill to remove globally redundant tokens while retaining sufficient visual coverage, and aggressive, query-aware retrieval is applied during decoding to focus computation on the most relevant visual cues. This adaptive allocation introduces sparsity where it yields the greatest efficiency gain, without compromising contextual grounding for future queries.

\begin{wraptable}{r}{0.43\textwidth}
    \vspace{-20pt}
    \small\centering
    \setlength{\tabcolsep}{4pt}
    \begin{tabular}{cccccc}
        \toprule
        \multicolumn{2}{c}{Sparsity} & \multicolumn{3}{c}{Speedup} & \multirow{2.5}{*}{\makecell{Robo \\ VQA}}\\
        \cmidrule(lr){1-2}\cmidrule(lr){3-5}
        Prefill & Decode & Prefill & Decode & E2E \\
        \midrule
        0\% & 0\% & 1.0$\times$ & 1.0$\times$ & 1.0$\times$ & 86.4 \\
        \cdashlinelr{1-6}
        \textbf{90\%} & 0\% & \textbf{14.6$\times$} & 1.1$\times$ & \textbf{1.4$\times$} & 80.0 \\
        \rowcolor{teal!15}
        70\% & \textbf{85\%} & 4.9$\times$ & \textbf{1.2$\times$} & \textbf{1.4$\times$} & \textbf{89.1} \\
        \bottomrule
    \end{tabular}
    \caption{Decoupled Prefill–Decode Sparsity}
    \label{tab:sparsity_ablation}
    \vspace{-20pt}
\end{wraptable}

We compare the decoupled design with a \textit{prefill-only} sparsity baseline on RoboVQA~\cite{sermanet2024robovqa} (\tab{tab:sparsity_ablation}). When tuned for equivalent end-to-end speedup, reallocating sparsity toward decoding consistently improves task performance. The prefill stage retains enough visual tokens to maintain context integrity, while decoding sparsity effectively targets the dominant latency source in multimodal generation.

\paragraph{Analysis.}
Retrieved tokens in \method exhibit two distinct functional roles: \textit{Visual Attention Sinks} and \textit{Visual Retrieval Tokens}. Sink tokens maintain stable activation across queries, acting as persistent attractors that stabilize cross-modal attention, whereas retrieval tokens vary dynamically with query content, capturing task-specific relevance. This separation explains how \method sustains contextual grounding while enabling efficient, query-adaptive retrieval. Qualitative analyses in Section~\ref{sec:qualitative} corroborate these behaviors, aligning with observations from VisionZip~\cite{yang2025visionzip} and Visual Attention Redistribution (VAR)~\cite{kang2025var}.

\section{Experiments}
\label{sec:experiments}

\subsection{Setup}

\paragraph{Multi-Turn Evaluation.} 

Many existing benchmarks generate multiple question–answer pairs for each image: \eg, up to 18 in POPE~\cite{li2023pope}. However, evaluation protocols often remain confined to single-turn settings. In this work, we make use of the inherent multi-turn structure of such datasets to enable more efficient and accurate assessment of VLMs. We organize questions associated with the same image into coherent multi-turn conversations, which allows visual tokens to be prefilled only once. This setup not only reduces computational overhead but also better reflects realistic VLM usage in interactive settings. However, a potential problem is information leakage between turns, where earlier questions can unintentionally reveal answers to later ones. For example, in the GQA dataset~\cite{hudson2019gqa}:
\vspace{2pt}
\begin{compactitem}
\item Q1: \textit{``What is the person in front of \textcolor{darkgreen}{the sky} doing?"}
\item Q2: \textit{``What's the person in front of?"}
\end{compactitem}
\vspace{2pt}
Here, Q1 discloses information that effectively answers Q2, enabling the model to respond correctly even without relying on the image. To mitigate this, we implement a partial KV cache eviction strategy: after each round, we remove only the KV entries corresponding to the previous question and answer. This preserves the efficiency gains of visual KV cache reuse while preventing unintended context carryover between turns. We will release our multi-turn evaluation framework to support the broader multimodal research community.

\paragraph{Baselines.}
We compare our \method with two categories of token pruning baselines: (i) \textit{query-agnostic methods}, which prune redundant visual tokens without reference to the textual input, including VisionZip~\cite{yang2025visionzip}, PruMerge~\cite{shang2025prumerge}, and HIRED~\cite{arif2025hired}; and (ii) \textit{query-aware methods}, which adapt token selection based on the language context, such as FastV~\cite{chen2024fastv}, PDrop~\cite{xing2025pyramiddrop}, and SparseVLM~\cite{zhang2025sparsevlm}. Most of these approaches estimate token salience using attention weights, which can be memory- and latency-intensive in visual encoders, often exceeding GPU limits at long context lengths. To ensure a fair comparison, we compute attention maps for these methods in a chunked manner to fit within memory constraints. In contrast, \method employs our fused Triton kernel to avoid materializing the full attention map (Figure~\ref{fig:triton_kernel}).

\paragraph{Inference Setting.}
We build an optimized inference pipeline based on TinyChat.
Specifically, we apply W8A8 quantization to the visual encoder following SmoothQuant~\cite{xiao2023smoothquant}, and W4A16 quantization to the LLM following AWQ~\cite{lin2024awq}.
This quantized version achieves a 2.4$\times$ end-to-end speedup over the vanilla one, with negligible accuracy degradation, as verified in preliminary experiments. 
All subsequent results in this work are reported on top of this quantized version. 
Unless otherwise stated, inference is performed on a single NVIDIA A6000 GPU using greedy decoding with a batch size of 1.

\paragraph{Latency Evaluation.}
We measure the end-to-end inference runtime, including the visual encoder (E), language model prefilling (P), and decoding throughput (D). 
Total latency (E2E) is defined as the sum of prefill time and per-token decoding time, with decoding lengths fixed to ensure consistency across tasks. 
Because our evaluation focuses on multi-turn conversations, we account for the chunked prefilling cost of queries across conversation rounds and amortize it into the initial image prefill stage. 
The number of rounds for each task is set to the average number of conversational turns observed in its dataset. 
For image-based tasks, we fix the decoding length to 50 tokens per round to emulate image captioning workloads. 
For video-based tasks, we use 250 tokens per round to approximate the latency of video captioning and detailed video generation. 
Reasoning models are evaluated in a single-turn setting, where total latency is computed as the sum of prefill and decoding times over 1,500 tokens, consistent with the typical 1-2K token output length of reasoning tasks.

\paragraph{Sparsity Ratio.}
Our sparsity configuration adopts a straightforward approach for both prefill and decoding, ensuring efficient implementation and cross-model compatibility. 
Specifically, we set a constant prefill sparsity before the LLM and a uniform decoding sparsity across all layers. 
More granular strategies, such as layer-wise or head-aware sparsity, may yield further optimization but introduce additional complexity and tuning overhead. 
We prioritize simplicity and generalization, leaving these refinements for future work.

\begin{table*}[t]
    \setlength{\tabcolsep}{2pt}
    \small\centering
    \begin{tabular}{lrrrrrrccccccccccc}
        \toprule
        & \multicolumn{2}{c}{Sparsity} & \multicolumn{4}{c}{Speedup} 
        & \multirow{2.5}{*}{AI2D} 
        & \multirow{2.5}{*}{\makecell{Chart \\ QA}} 
        & \multirow{2.5}{*}{\makecell{Doc \\ VQA}} 
        & \multirow{2.5}{*}{GQA} 
        & \multirow{2.5}{*}{\makecell{Info \\ VQA}} 
        & \multirow{2.5}{*}{\makecell{MME \\ Sum}} 
        & \multirow{2.5}{*}{POPE} 
        & \multirow{2.5}{*}{SQA} 
        & \multirow{2.5}{*}{\makecell{Text \\ VQA}} \\
         \cmidrule(lr){2-3}\cmidrule(lr){4-7}
          & P & D & E & P & D & E2E \\
          \midrule
        LLaVA-NeXT-7B & 0 & 0 & 1.0$\times$ & 1.0$\times$ & 1.0$\times$ & 1.0$\times$ 
            & 63.9 & 53.0 & 63.6 & 63.5 & 28.4 & 1857.8 
            & 84.5 & 69.3 & 58.2 \\
        \cdashlinelr{1-16}
        ~~+ FastV & .80 & 0 & 1.0$\times$ & 1.5$\times$ & 1.2$\times$ & 1.2$\times$
            & 61.8 & 31.6 & 33.5 & 55.3 & 22.0 & 1568.2 
            & 76.7 & 66.7 & 52.7 \\
        ~~+ SparseVLM & .75 & 0 & 1.0$\times$ & 1.4$\times$ & 1.2$\times$ & 1.2$\times$
            & 63.2 & 39.9 & 41.8 & 59.7 & 22.2 & 1823.9 
            & 83.4 & \textbf{69.6} & 57.6 \\
        ~~+ PDrop & .64 & 0 & 1.0$\times$ & 1.3$\times$ & 1.2$\times$ & 1.2$\times$ 
            & 62.9 & 33.6 & 25.3 & 54.6 & 20.4 & 1793.4 
            & 81.6 & 68.9 & 52.6 \\
        ~~+ PruMerge & .80 & 0 & 0.3$\times$ & 1.5$\times$ & 1.2$\times$ & 1.2$\times$
            & 60.0 & 25.4 & 26.9 & 59.8 & 21.4 & 1686.2 
            & 82.0 & 67.8 & 48.3 \\
        ~~+ HIRED & .80 & 0 & 0.9$\times$ & 1.5$\times$ & 1.2$\times$ & 1.2$\times$ & 59.4 & 33.4 & 34.8 & 60.4 & 22.0 & 1560.0 & 80.5 & 68.7 & 50.6 \\
        ~~+ VisionZip & .80 & 0 & 1.0$\times$ & 1.5$\times$ & 1.2$\times$ & 1.2$\times$
            & 62.9 & 38.2 & 48.5 & 60.3 & 24.2 & 1727.4 
            & 84.1 & 67.9 & 57.1 \\
        ~~+ \textbf{\method} & .60 & .75 & 1.0$\times$ & 1.1$\times$ & 1.2$\times$ & 1.2$\times$
            & \textbf{64.1} & \textbf{47.8} & \textbf{58.0} & \textbf{62.7} & \textbf{25.6} & \textbf{1831.0} 
            & \textbf{85.8} & \textbf{69.6} & \textbf{59.1} \\
        \midrule
        InternVL3.5-8B & 0 & 0 & 1.0$\times$ & 1.0$\times$ & 1.0$\times$ & 1.0$\times$ 
        & 80.9 & 79.2 & 86.4 & 60.5 & 72.8 & 2309.4 
        & 88.1 & 96.8 & 75.3\\
        \cdashlinelr{1-16}
        ~~+ FastV & .80 & 0 & 1.0$\times$ & 2.6$\times$ & 1.4$\times$ & 1.6$\times$ & 67.8 & 28.4 & 23.7 & 48.3 & 26.9 & 1909.4 & 72.5 & 80.5 & 54.8 \\
        ~~+ \textbf{\method} & .45 & .95 & 1.0$\times$ & 1.5$\times$ & 1.6$\times$ & 1.6$\times$ & \textbf{77.0} & \textbf{61.8} & \textbf{56.4} & \textbf{58.9} & \textbf{58.1} & \textbf{2276.3} & \textbf{87.9} & \textbf{94.3} & \textbf{67.5}  \\
        \midrule
        Nemotron-Nano-VL-8B & 0 & 0 & 1.0$\times$ & 1.0$\times$ & 1.0$\times$ & 1.0$\times$ 
            & 82.2 & 86.3 & 90.6 & 64.6 & 76.1 & 1936.4 
            & 87.7 & 97.2 & 82.9 \\
        \cdashlinelr{1-17}
        ~~+ \textbf{\method} & .60 & .75 & 1.0$\times$ & 1.1$\times$ & 1.3$\times$ & 1.3$\times$ & \textbf{79.7} & \textbf{66.1} & \textbf{71.7} & \textbf{63.6} & \textbf{60.1}& \textbf{1859.8} & \textbf{87.4} & \textbf{96.2} & \textbf{71.2} \\
        \bottomrule
    \end{tabular}
    \caption{\textbf{Image Benchmark Results}. \method preserves near-lossless accuracy on general VQA tasks and reduces degradation on document and chart benchmarks by a large margin compared to prior pruning methods, demonstrating stronger retention of fine-grained visual details.}
    \label{tab:results:image}
\end{table*}

\subsection{Image Benchmark Results}
\label{sec:experiments:image_benchmarks}

We evaluate \method across nine vision-language benchmarks, including AI2D~\cite{kembhavi2016ai2d}, ChartQA~\cite{masry2022chartqa}, DocVQA~\cite{mathew2021docvqa}, GQA~\cite{hudson2019gqa}, InfoVQA~\cite{mathew2022infographicvqa}, MME~\cite{fu2023mme}, 
POPE~\cite{li2023pope}, ScienceQA~\cite{lu2022scienceqa}, and TextVQA~\cite{singh2019textvqa}. These benchmarks span diagram understanding, document reasoning, and general visual question answering, enabling a comprehensive evaluation of efficiency-accuracy trade-offs.

As shown in \tab{tab:results:image}, when applied to LLaVA-NeXT-7B~\cite{liu2024llavanext}, \method maintains near-lossless performance on general VQA tasks such as AI2D, GQA, POPE, and ScienceQA, achieving accuracy comparable to or even higher than the unpruned baseline under high sparsity. On fine-grained document and chart understanding benchmarks (ChartQA, DocVQA, and InfoVQA), \method exhibits over \textbf{15\% less degradation} than prior pruning and merging methods such as FastV, SparseVLM, and VisionZip. These improvements stem from the decoupled sparsity design, which balances lightweight prefill pruning with adaptive decoding retrieval, preserving essential visual context across diverse vision-language tasks.

In addition to LLaVA-NeXT-7B, we also evaluate \method on InternVL3.5-8B~\cite{wang2025internvl35} and Llama-Nemotron-Nano-VL-8B~\cite{nvidia2025nemotronnanovl}. Across both models, \method maintains competitive accuracy while accelerating inference, demonstrating the generality of our decoupled sparsity framework across architectures and tasks.

\subsection{Video Benchmark Results}
\label{sec:experiments:video_benchmarks}

We then evaluate \method on diverse video benchmarks covering video question answering, captioning, and retrieval. These tasks assess the model's ability to handle extended visual contexts, sustain coherent generation over long sequences, and preserve fine-grained visual memory across multi-turn interactions. Across all benchmarks, \method delivers consistent improvements in both efficiency and accuracy by decoupling sparsity between the prefill and decoding stages. By shifting sparsity toward decoding, where query-aware retrieval selects only the most relevant visual tokens from the cached context, \method achieves faster throughput and stronger long-context retention than prior pruning-based methods, while maintaining fidelity in temporal understanding and generation quality.

\subsubsection{Video Understanding}  

We evaluate \method on four long-context video understanding benchmarks: LongVideoBench~\cite{wu2024longvideobench}, MLVU~\cite{zhou2025mlvu}, NExT-QA~\cite{xiao2021nextqa}, and Video-MME~\cite{fu2025videomme}. As shown in \tab{tab:results:video-qa}, \method consistently outperforms baselines across models~\cite{chen2025longvila,bai2025qwen25vl,nvidia2025nemotronnanovl}. Query-aware methods such as FastV, SparseVLM, and PDrop fail to scale beyond 32 frames due to their reliance on full joint query–vision attention, while query-agnostic methods like VisionZip and PruMerge introduce substantial overhead from token clustering and merging, which can even slow inference despite token reduction. In contrast, \method's decoupled sparsity framework scales efficiently to long video contexts by combining query-agnostic pruning during prefill with query-aware retrieval during decoding. This design achieves up to \textbf{6.0$\times$} faster language model prefill, \textbf{2.5$\times$} faster decoding, and an overall \textbf{2.6$\times$} end-to-end speedup, while maintaining near-lossless accuracy across all video understanding benchmarks.

Unlike the image benchmarks, \method even improves accuracy over the unpruned baseline on video benchmarks. We attribute this to more precise token retrieval enabled by a compact and information-dense KV cache, which helps the model focus on the most relevant visual cues. This also aligns with findings from StreamingLLM~\cite{xiao2024streamingllm}, where smaller active contexts were shown to improve focus and reasoning. Overall, these results indicate that decoupling sparsity not only enhances efficiency but also sharpens the model's attention to semantically important information, improving both accuracy and scalability in long-context video understanding.

\begin{table*}[t]
    \small\centering
    \setlength{\tabcolsep}{2pt}
    \begin{tabular}{lrrrrrrccccccc}
        \toprule
        & \multicolumn{2}{c}{Sparsity} & \multicolumn{4}{c}{Speedup} & LVB & MLVU & NExT-QA & \multicolumn{4}{c}{Video-MME (w/o sub)} \\
        \cmidrule(lr){2-3}\cmidrule(lr){4-7}\cmidrule(lr){8-8}\cmidrule(lr){9-9}\cmidrule(lr){10-10}\cmidrule(lr){11-14}
        & P & D & E & P & D & E2E & val & m-avg & mc & S & M & L & Overall \\
        \midrule
        LongVILA-7B \textcolor{gray}{(256f)} & 0 & 0 & 1.0$\times$ & 1.0$\times$ & 1.0$\times$ & 1.0$\times$ & 53.8 & 64.9 & 78.6 & 67.6 & 57.7 & 51.2 & 58.8 \\
        \cdashlinelr{1-14}
        ~~+ VisionZip & .95 & 0 & 0.9$\times$ & 28.5$\times$ & 1.5$\times$ & 2.1$\times$ & 47.0 & 60.4 & 75.5 & 58.0 & 51.6 & 47.0 & 52.2 \\
        ~~+ PruMerge & .95 & 0 & 0.9$\times$ & 28.5$\times$ & 1.5$\times$ & 2.1$\times$ & 47.9 & 60.9 & 75.7 & 57.9 & 51.6 & 46.7 & 52.0 \\
        ~~+ \textbf{\method} & .75 & .90 & 1.0$\times$ & 5.1$\times$ & 1.6$\times$ & 2.1$\times$ & \textbf{54.1} & \textbf{65.3} & \textbf{79.0} & \textbf{68.3} & \textbf{58.2} & \textbf{49.6} & \textbf{58.7} \\
        \midrule
        Qwen2.5-VL-7B \textcolor{gray}{(4fps)} & 0 & 0 & 1.0$\times$ & 1.0$\times$ & 1.0$\times$ & 1.0$\times$ & 59.2 & 65.5 & 76.0 & 73.0 & 60.8 & 53.1 & 62.3 \\
        \cdashlinelr{1-14}
        ~~+ \textbf{\method} & .75 & .90 & 0.4$\times$ & 6.0$\times$ & 2.0$\times$ & 1.9$\times$ & \textbf{60.1} & \textbf{70.7} & \textbf{81.9} & \textbf{75.9} & \textbf{65.9} & \textbf{57.1} & \textbf{66.3} \\
        \midrule
        Nemotron-Nano-VL-8B \textcolor{gray}{(256f)} & 0 & 0 & 1.0$\times$ & 1.0$\times$ & 1.0$\times$ & 1.0$\times$ & 55.3 & 60.9 & 75.8 & 68.3 & 51.6 & 45.8 & 55.2 \\
        \cdashlinelr{1-14}
        ~~+ \textbf{\method} & .75 & .95 & 1.0$\times$ & 4.0$\times$ & 2.5$\times$ & 2.6$\times$ & \textbf{55.9} & \textbf{63.1} & \textbf{76.6} & \textbf{68.9} & \textbf{54.2} & \textbf{46.8} & \textbf{56.6} \\
        \bottomrule
    \end{tabular}
    \caption{\textbf{Video Understanding Benchmark Results}. \method delivers up to \textbf{6.0$\times$} faster language model prefill, \textbf{2.5$\times$} faster decoding, and \textbf{2.6$\times$} end-to-end speedup, maintaining \textbf{near-lossless} accuracy on video understanding.}
    \label{tab:results:video-qa}
\end{table*}

\begin{table}[t]
    \small\centering
    \begin{tabular}{lrrrrrrcccccc}
        \toprule
        & \multicolumn{2}{c}{Sparsity} & \multicolumn{4}{c}{Speedup} & \multicolumn{6}{c}{Video-ChatGPT} \\
        \cmidrule(lr){2-3}\cmidrule(lr){4-7}\cmidrule(lr){8-13}
        & P & D & E & P & D & E2E & CI & DO & CU & TU & C & Overall \\
        \midrule
        LongVILA-7B \textcolor{gray}{(256f)} & 0 & 0 & 1.0$\times$ & 1.0$\times$ & 1.0$\times$ & 1.0$\times$ & 2.34 & 2.21 & 2.81 & 1.70 & 2.46 & 2.31 \\
        \cdashlinelr{1-13}
        ~~+ VisionZip & .95 & 0 & 0.9$\times$ & 28.5$\times$ & 1.5$\times$ & 2.1$\times$ & 2.04 & 2.03 & 2.56 & 1.71 & 2.11 & 2.09 \\
        ~~+ PruMerge & .95 & 0 & 0.9$\times$ & 28.5$\times$ & 1.5$\times$ & 2.1$\times$ & 2.07 & 2.00 & 2.57 & 1.75 & 2.10 & 2.10\\
        ~~+ \textbf{\method} & .75 & .90 & 1.0$\times$ & 5.1$\times$ & 1.6$\times$ & 2.1$\times$ & \textbf{2.35} & \textbf{2.27} & \textbf{2.85} & \textbf{1.90} & \textbf{2.39} & \textbf{2.35} \\
        \bottomrule
    \end{tabular}
    \caption{\textbf{Video Captioning Benchmark Results}. \method delivers \textbf{5.1$\times$} faster prefill, \textbf{1.6$\times$} faster decoding and \textbf{2.1$\times$} end-to-end speedup while slightly improving the overall Video-ChatGPT score. It delivers consistent gains across \textit{correctness} (CI), \textit{detail} (DO), \textit{contextual understanding} (CU), \textit{temporal understanding} (TU), and \textit{consistency} (C), outperforming baselines with more coherent video generation. Evaluation scores are obtained using \texttt{gpt-4o-mini-2024-07-18}.}
    \label{tab:results:video-cap}
\end{table}

\subsubsection{Video Captioning}

We further evaluate \method on long-generation tasks using the VideoChatGPT benchmark~\cite{maaz2023videochatgpt}, which measures a model's ability to produce extended, free-form video descriptions under long-context settings. This benchmark provides a \texttt{GPT}-aided evaluation across five dimensions: \textit{correctness}, \textit{detail}, \textit{contextual understanding}, \textit{temporal understanding}, and \textit{consistency}. 

As shown in \tab{tab:results:video-cap}, \method maintains high generation quality while achieving substantial computational savings. Compared to the unpruned baseline, it improves the overall Video-ChatGPT score from \textbf{2.31} to \textbf{2.35}, with a \textbf{0.2} gain in temporal understanding. Unlike VisionZip and PruMerge, which both suffer moderate degradation in contextual understanding, \method preserves coherence and factual grounding even under high decoding sparsity. It achieves up to a \textbf{2.1$\times$} end-to-end speedup, demonstrating that query-aware retrieval during decoding effectively maintains semantic grounding and enhances detail richness, while directly addressing the dominant latency bottleneck in multimodal generation.

\begin{figure*}[t]
    \centering
    \includegraphics[width=\linewidth]{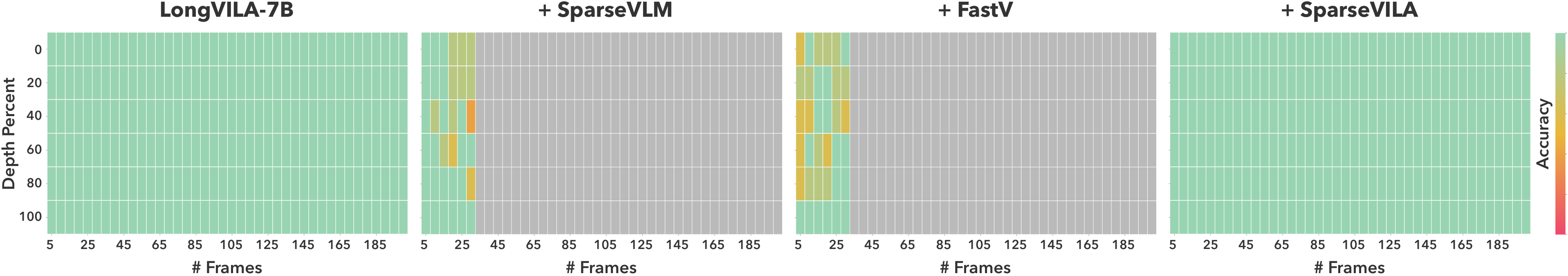}
    \caption{\textbf{Visual Retrieval Results}. SparseVLM and FastV degrade and fail beyond 32 frames (8K context), while \method maintains perfect retrieval up to 200 frames, demonstrating superior long-context scalability.}
    \label{fig:longctx-vniah}
\end{figure*}

\subsubsection{Visual Retrieval}

We evaluate long-context retrieval performance on a multi-turn Visual Needle-in-a-Haystack (V-NIAH) benchmark, extended from LongVILA~\cite{chen2025longvila} and LongVA~\cite{zhang2024longva}. Each sequence contains five target ``needles'' interleaved among haystack frames. To emulate realistic conversational interaction, all needles are embedded within the haystack, and the model is prompted sequentially with one of the five corresponding queries, while the remaining needles serve as distractors. This design ensures that accurate retrieval depends on identifying the correct visual segment rather than leveraging correlations among other embedded needles. To account for depth sensitivity, we re-prefill the context at each depth by conditioning on the first query, reapplying pruning, and then issuing the target question for evaluation.

Using LongVILA-7B~\cite{chen2025longvila}, we compare our \method with SparseVLM and FastV across progressively longer visual contexts. Both SparseVLM and FastV fail to scale beyond 32 frames due to their reliance on joint query–vision attention, which results in excessive memory consumption. Even within this range, they show early degradation in retrieval accuracy as context length increases. In contrast, \method sustains near-perfect retrieval up to 200 frames, demonstrating strong long-context retention (see \fig{fig:longctx-vniah}). This robustness stems from \method's query-aware decoding sparsity, which selectively retrieves relevant visual cues from the preserved KV cache instead of pruning them during the prefill stage.

\subsection{Reasoning Benchmark Results}
\label{sec:experiments:reasoning}

We evaluate \method on long-context and physical reasoning workloads that stress multimodal inference beyond standard VQA or captioning. These tasks typically require substantially longer generations, making decoding throughput the dominant contributor to end-to-end latency. Prior sparsity methods concentrate compression in the prefill stage and thus provide limited benefit in this regime. In contrast, \method's decoupled design allocates lightweight, query-agnostic pruning to prefill while shifting aggressive, query-aware retrieval to decoding, preserving reasoning fidelity under long outputs and delivering practical speedups.

\subsubsection{Video Reasoning}

We assess \method on LongVideo-Reason~\cite{chen2025longvilar1}, which features complex question–answer pairs requiring temporal reasoning over extended video sequences. As reported in \tab{tab:longvila_r1}, \method consistently outperforms state-of-the-art pruning and merging approaches (\eg, PruMerge, VisionZip) while achieving up to \textbf{1.3$\times$} faster inference. The gains stem from reallocating sparsity toward decoding, where query-aware retrieval narrows attention to the most relevant visual tokens in the cached context without discarding information needed for later turns. This maintains long-horizon temporal consistency and yields higher answer accuracy at comparable end-to-end speedups.

\subsubsection{Physical Reasoning}
We evaluate \method on physical reasoning suites that demand causal understanding and multi-step deduction. As shown in \tab{tab:physical_reasoning}, \method matches or exceeds baselines across all tasks, operating on the Pareto frontier of efficiency and accuracy. Notably, \method surpasses the unpruned model on all subsets at 24 frames-per-second while delivering a lossless \textbf{1.9$\times$} end-to-end speedup and a \textbf{4.5\%} performance gain. These results indicate that concentrating sparsity in decoding sharpens the model’s focus on semantically critical evidence, preserving structured reasoning under aggressive compression.

\begin{figure*}[t]
    \centering
    \includegraphics[width=\linewidth]{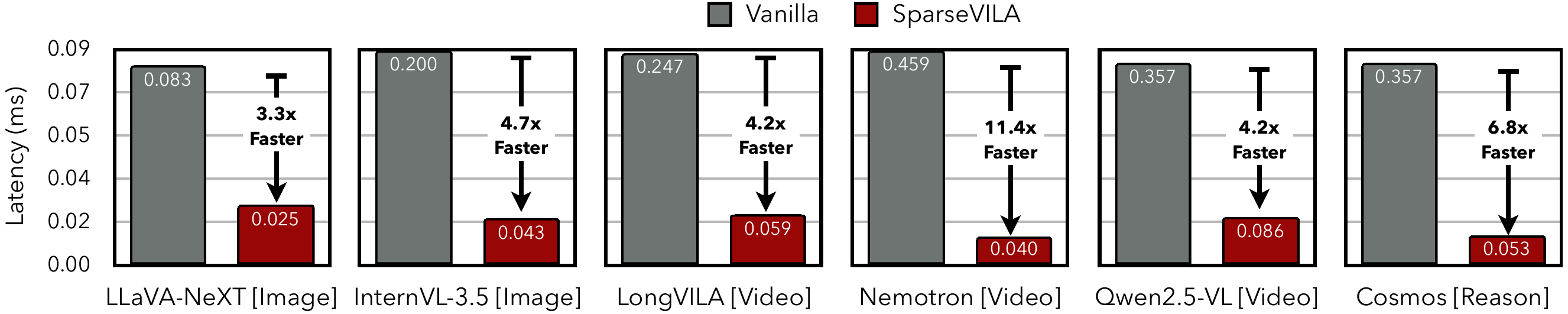}
    \caption{\textbf{Decoding Attention Kernel.} \method's decoupled sparsity reduces the size of the effective KV cache in decoding, delivering up to \textbf{11.4$\times$} speedup on the attention kernel for long video understanding tasks.}
    \label{fig:latency-comparison}
\end{figure*}

\paragraph{Discussion.}
Across both video and physical reasoning settings, many pruning methods reduce theoretical compute yet incur substantial overhead from salience computation or token reorganization, limiting realized speedups. In contrast, \method combines stage-aware sparsity with fused Triton kernels (\fig{fig:latency-comparison}), keeping overhead low and aligning empirical latency with theoretical gains. The decoupled design retains a rich visual KV cache for future turns while activating only the query-relevant subset during generation, yielding robust accuracy and consistent acceleration in reasoning-heavy workloads.

\begin{table*}[t]
    \setlength{\tabcolsep}{5.5pt}
    \small\centering
    \begin{tabular}{lrrrrrrccccc}
        \toprule
        & \multicolumn{2}{c}{Sparsity} & \multicolumn{4}{c}{Speedup} 
        & \multirow{2.5}{*}{Temporal} & \multirow{2.5}{*}{Goal} & \multirow{2.5}{*}{Plot} 
        & \multirow{2.5}{*}{Spatial} & \multirow{2.5}{*}{Overall} \\
         \cmidrule(lr){2-3}\cmidrule(lr){4-7}
          & P & D & E & P & D & E2E \\
        \midrule
        LongVILA-R1-7B \textcolor{gray}{(512f)} & 0 & 0 & 1.0$\times$ & 1.0$\times$ & 1.0$\times$ & 1.0$\times$ & 75.5 & 88.9 & 74.9 & 58.5 & 74.5 \\
        \cdashlinelr{1-12}
        ~~+ PruMerge & .95 & 0 & 0.8$\times$ & 10.5$\times$ & 1.1$\times$ & 1.1$\times$ & 63.3 & 86.5 & \textbf{73.9} & 57.3 & 72.9 \\
        ~~+ VisionZip & .95 & 0 & 0.9$\times$ & 10.5$\times$ & 1.1$\times$ & 1.2$\times$ & 62.9 & 85.4 & 70.3 & \textbf{61.0} & 71.8 \\
        ~~+ \textbf{\method} & .75 & .90 & 1.0$\times$ &  4.5$\times$ & 1.2$\times$ & 1.3$\times$ & \textbf{66.7} & \textbf{87.8} & 73.1 & 59.8 & \textbf{74.4} \\
        \bottomrule
    \end{tabular}
    \caption{\textbf{Video Reasoning Benchmark Results}.  \method maintains competitive performance on long-video reasoning tasks while delivering up to  \textbf{1.3$\times$} end-to-end speedup.}
    \label{tab:longvila_r1}
\end{table*}

\begin{table*}[t]
    \setlength{\tabcolsep}{4pt}
    \small\centering
    \begin{tabular}{lrrrrrrcccc}
        \toprule
        & \multicolumn{2}{c}{Sparsity} & \multicolumn{4}{c}{Speedup} 
        & \multirow{2.5}{*}{HoloAssist} & \multirow{2.5}{*}{RoboFail} 
        & \multirow{2.5}{*}{RoboVQA} & \multirow{2.5}{*}{Average} \\
        \cmidrule(lr){2-3}\cmidrule(lr){4-7}
        & P & D & E & P & D & E2E \\
        \midrule
        Cosmos-Reason1-7B \textcolor{gray}{(4fps)} & 0 & 0 & 1.0$\times$ & 1.0$\times$ & 1.0$\times$ & 1.0$\times$ & 65.0 & 60.0 & 86.4 & 70.5 \\
        \cdashlinelr{1-11}
        ~~+ PruMerge & .90 & 0 & 0.2$\times$ & 2.2$\times$ & 1.1$\times$ & 0.7$\times$ & 41.0 & 39.0 & 52.3 & 44.2\\
        ~~+ VisionZip & .90 & 0 & 0.2$\times$ & 14.6$\times$ & 1.1$\times$ & 0.8$\times$ & \textbf{66.0} & 54.0 & 80.3 & 66.7\\
        ~~+ FastV & .71 & 0 & 1.0$\times$ & 2.2$\times$ & 1.1$\times$ & 1.3$\times$ & 46.0 & 37.0 & 80.9 & 52.5 \\
        ~~+ \textbf{\method} & .70 & .85 & 0.7$\times$ & 4.9$\times$ & 1.2$\times$ & 1.4$\times$ & 64.0 & \textbf{63.0} & \textbf{89.1} & \textbf{72.0} \\
        \midrule
        Cosmos-Reason1-7B \textcolor{gray}{(24fps)} & 0 & 0 & 1.0$\times$ & 1.0$\times$ & 1.0$\times$ & 1.0$\times$ & 72.0 & 54.0 & 88.2 & 71.4 \\
        \cdashlinelr{1-11}
        ~~+ PruMerge & .97 & 0 & 0.04$\times$ & 13.8$\times$ & 1.1$\times$ & 0.7$\times$ & 46.0 & 43.0 & 70.0 & 53.0\\
        ~~+ VisionZip & .97 & 0 & 0.04$\times$ & 73.4$\times$ & 1.6$\times$ & 0.3$\times$ & 64.0 & 54.0 & 80.9 & 66.6 \\
        ~~+ \textbf{\method} & .75 & .95 & 0.4$\times$ & 7.6$\times$ & 2.0$\times$ & 1.9$\times$ & \textbf{75.0} & \textbf{58.0} & \textbf{94.5} & \textbf{75.9} \\
        \bottomrule
    \end{tabular}
    \caption{\textbf{Physical Reasoning Benchmark Results}. \method delivers up to \textbf{7.6$\times$} faster language model prefill, \textbf{2.0$\times$} faster decoding, and \textbf{1.9$\times$} end-to-end speedup, while outperforming prior methods and the baseline model at 24 frames-per-second.}
    \label{tab:physical_reasoning}
\end{table*}

\subsection{Efficiency Analysis}
\label{sec:experiments:efficiency}

\method achieves consistent acceleration across image, video, and reasoning workloads through its decoupled sparsity framework. This design scales effectively across diverse architectures and attention mechanisms, including standard multi-head attention (MHA) in LLaVA-NeXT~\cite{liu2024llavanext} and grouped-query attention (GQA) in LongVILA-7B~\cite{chen2025longvila}, Qwen2.5VL-7B~\cite{yang2024qwen2}, and InternVL3.5-8B~\cite{wang2025internvl35}. By combining custom kernel design with stage-aware sparsity allocation, \method provides a full-stack optimization of the VLM inference pipeline, spanning embedding, prefill, and decoding. This holistic design captures both compute- and memory-bound stages, improving end-to-end latency while maintaining fidelity. Furthermore, \method’s decoupled sparsity scales seamlessly from short-context image tasks to long-horizon video and reasoning settings, ensuring stable efficiency across heterogeneous model families. Besides, \method's prefill-stage pruning provides complementary memory savings, reducing KV memory usage by \textbf{72.5\%} and linear FLOPS by \textbf{87.6\%} on LongVILA-7B through structured token sparsity.

\paragraph{Decoding Attention Kernel Efficiency.}
We further analyze performance at the kernel level to isolate the contribution of our optimization from model-level sparsity. The decoding stage of LLM inference is memory-bound and thus bottlenecked by memory movement. By reducing the effective size of the KV Cache, \method lowers both memory traffic and decoding FLOPs, leading to substantial acceleration. As shown in Figure~\ref{fig:latency-comparison}, \method delivers up to \textbf{11.4$\times$} speedup on long-context video workloads and \textbf{6.8$\times$} on reasoning tasks.

\paragraph{Empirical vs. Theoretical Latency Analysis}
A key factor in evaluating sparsity strategies is the gap between theoretical and realized latency. Even when token reduction establishes a clear upper bound on achievable speedup, additional computation can diminish these gains in practice. The reported latency measurements therefore capture both the benefits of sparsity and the method-specific overhead incurred during inference. This overhead arises from pruning metric computation, token reorganization, and selection logic. Query-aware methods, which delay pruning to deeper layers, introduce nontrivial computational cost. VisionZip~\cite{yang2025visionzip} exhibits significant overhead in the embedding stage due to full attention weight computation, limiting effective speedup at long contexts. Similarly, PruMerge~\cite{shang2025prumerge} incurs additional prefill-stage overhead due to clustering-based pruning. In contrast, \method maintains low overhead in both prefill and decoding, resulting in empirical latency that more closely aligns with the theoretical sparsity bound. Table~\ref{tab:sota:overheads} summarizes the measured CUDA-time overhead for select methods across video and reasoning workloads.

\begin{table}[t]
    \small\centering
\setlength{\tabcolsep}{4pt}
\begin{tabular}{lcccccccc}
\toprule
\multirow{2}{*}{Method} &
\multicolumn{4}{c}{LongVILA-7B \textcolor{gray}{(256f)}} &
\multicolumn{4}{c}{Cosmos-Reason1-7B \textcolor{gray}{(4fps)}} \\
& \multicolumn{2}{c}{Overhead} & \multicolumn{2}{c}{Speedup} & \multicolumn{2}{c}{Overhead} & \multicolumn{2}{c}{Speedup} \\
\cmidrule(lr){2-3}\cmidrule(lr){4-5}
\cmidrule(lr){6-7}\cmidrule(lr){8-9}
 & CUDA (ms)& \% & T & E & CUDA (ms)& \% & T & E \\
\midrule
PruMerge        &  448.3 & 1.84 & 7.41$\times$ & 6.49$\times$ & 20990.1 & 171.4 & 3.04$\times$ & 0.49$\times$\\
VisionZip       &  206.3 & 0.85 & 7.41$\times$ & 6.94$\times$ & 17612.3 & 143.8 & 3.04$\times$ & 0.57$\times$\\
\method &  94.9 & 0.39 & 3.98$\times$ & 3.91$\times$ & 1400.9 & 26.1 & 2.28$\times$ & 1.81$\times$\\
\bottomrule
\end{tabular}
    \caption{\textbf{Overhead comparison across workloads.} Comparison of the overhead incurred by different methods on LongVILA-7B (video), and Cosmos-Reason1-7B (reasoning) workloads. CUDA Time (ms) denotes the additional latency measured on a single NVIDIA A6000 GPU, and the percentage indicates the relative overhead during the media-prefilling stage. We additionally report the theoretical (T) and empirical (E) speedups for each model/workload corresponding to settings in Section~\ref{sec:experiments}.}
    \label{tab:sota:overheads}
\end{table}

\subsection{Qualitative Analysis}
\label{sec:qualitative}

\begin{figure*}[t]
    \centering
    \includegraphics[width=\linewidth]{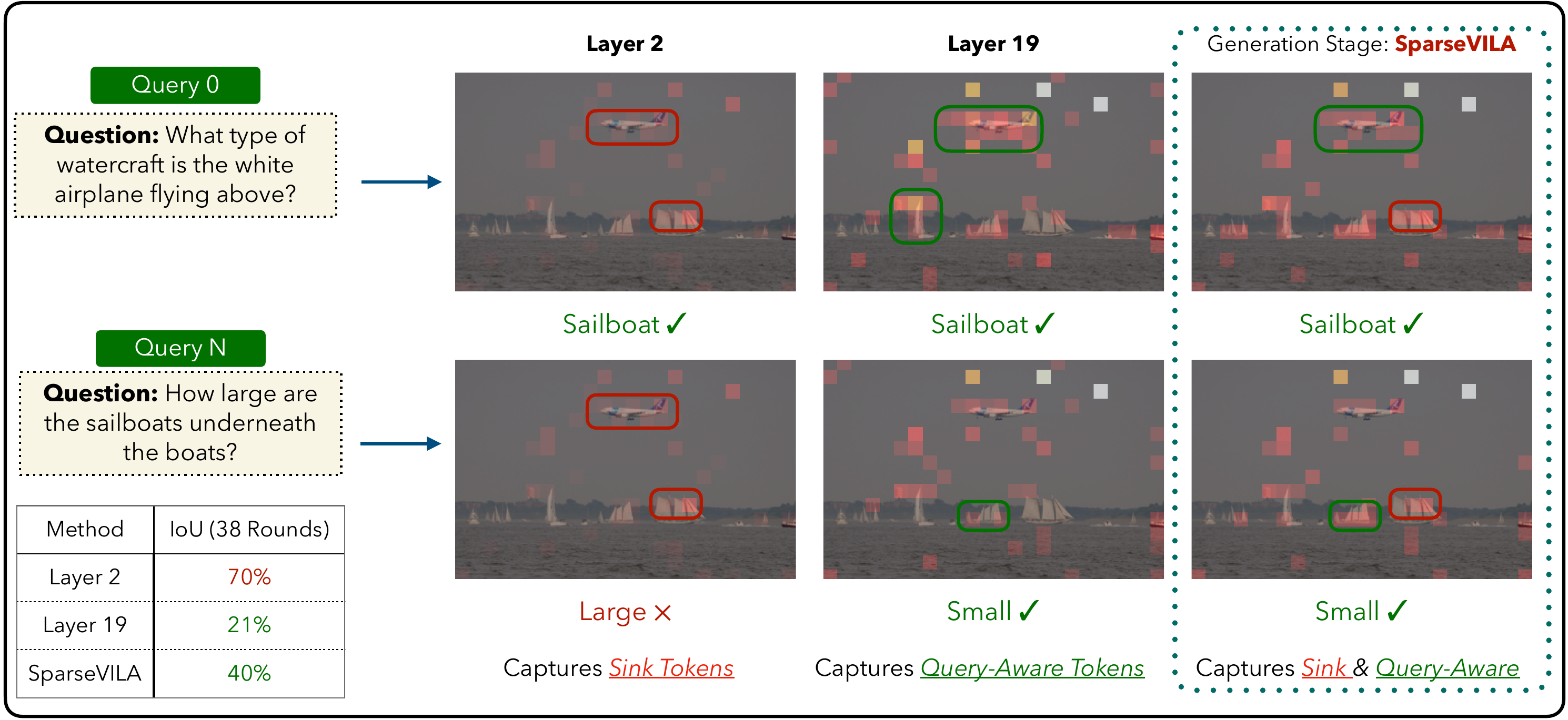}
    \caption{\textbf{Emergence of sink and retrieval tokens.}
    Early layers emphasize persistent visual sinks, while deeper layers highlight query-dependent retrieval tokens. This observation motivates \method's decoupled design, which preserves both phenomena through full-context attention before sparsification.}
    \label{fig:sink_retrieval}
\end{figure*}

\paragraph{Emergence of sink and retrieval tokens.}
In Figure~\ref{fig:sink_retrieval}, we examine how visual \textit{sink} and \textit{retrieval} tokens emerge throughout the LLM layers of LLaVA-1.5. Profiling the attention maps reveals that early layers concentrate on a small subset of visual tokens that remain stable across different queries -- these correspond to persistent \textit{visual sink tokens} that act as anchors of scene understanding. As depth increases, attention patterns diversify and \textit{retrieval tokens} emerge, focusing selectively on regions relevant to the query. The sink tokens continue to exist but with diminished strength, indicating that query-specific reasoning gradually overrides the globally salient structures. Quantitatively, we can use the intersection-over-union (IoU) of selected tokens for different input queries to quantify the proportion of sink and retrieval tokens captured. On the first 38 multi-turn queries in the GQA dataset, the IoU is highest in shallow layers (e.g., Layer~2), confirming strong sink consistency over different queries, and decreases toward deeper layers (e.g., Layer~19), where query-dependent retrieval dominates.

\paragraph{Design implications for \method.}
Considering that retrieval tokens only emerge in deeper layers, pruning tokens early in the network -- as done by many prefill-only methods -- irreversibly removes information essential for query-aware reasoning. \method, therefore, is specifically designed to preserve maximum content in the prefill phase, thereby deferring the bulk of pruning decisions into the decoding stage. It then performs selection across all layers, ensuring that both persistent sinks and query-dependent retrievals are retained. By shifting aggressive sparsity into the decoding stage, \method exploits this separation to balance compression and contextual fidelity, preserving long-term grounding without redundant visual computation.

\begin{figure*}[t]
    \centering
    \includegraphics[width=\linewidth]{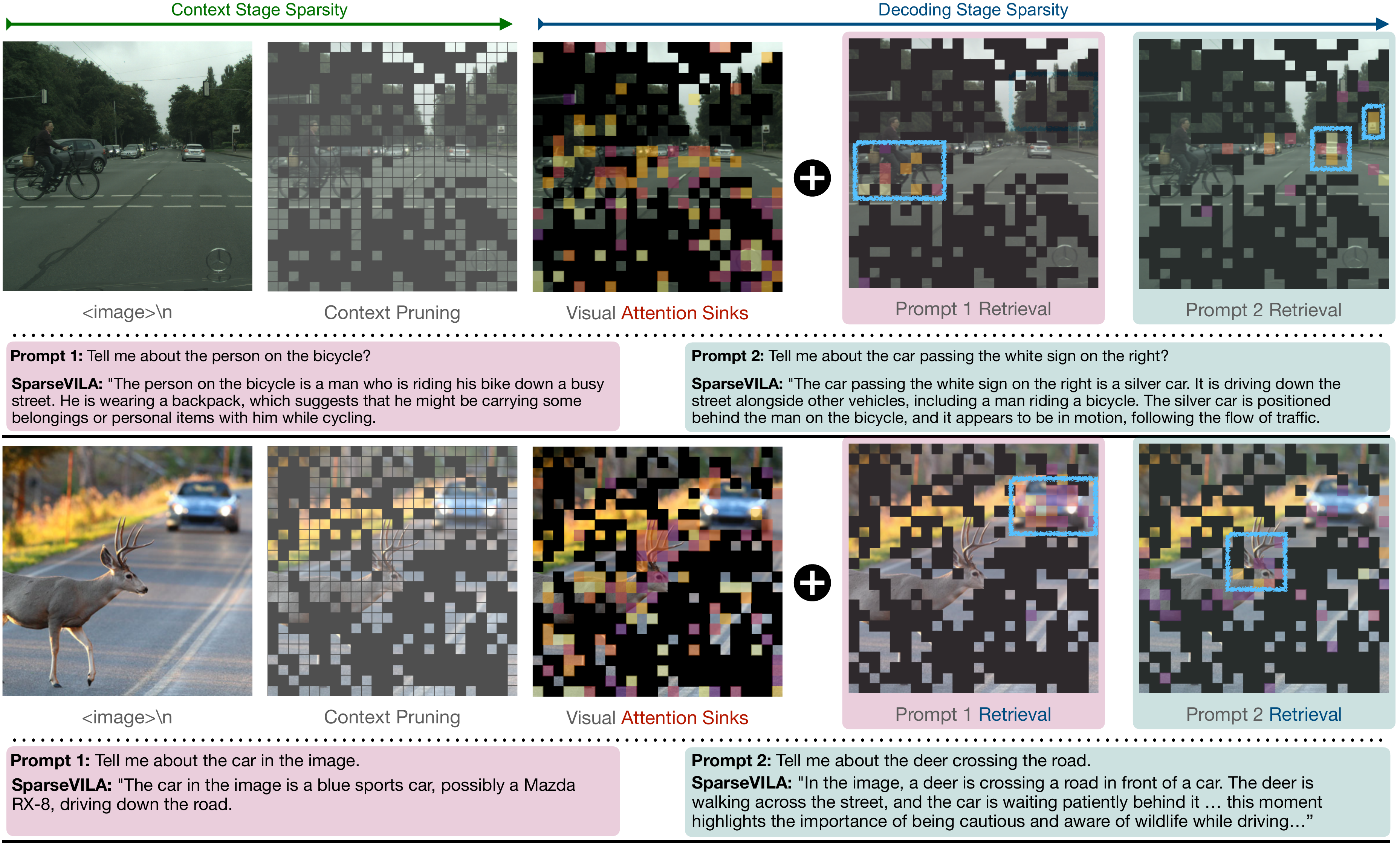}
    \caption{\textbf{Token selection frequency under 50\% context and 75\% decoding sparsity.}
    Each heatmap shows how often a visual token is selected across all LLM layers for a given query on LLaVA-1.5. \method maintains stable sink tokens while retrieving query-specific tokens, achieving efficient yet faithful decoding.}
    \label{fig:supp:token_visual}
\end{figure*}

\paragraph{\method token selection.}
To visualize this effect, Figure~\ref{fig:supp:token_visual} shows token selection frequencies under 50\% context sparsity and 75\% decoding sparsity. Each heatmap depicts how often a visual token is chosen across all LLM layers for a given query. \method consistently preserves sink tokens -- regions repeatedly selected across layers -- while dynamically retrieving query-specific tokens corresponding to objects of interest. As illustrated in the examples, retrieval focuses shift appropriately across questions (e.g., the cyclist vs.\ the signpost), demonstrating that the decoupled design maintains both stability and adaptability in token selection.

\section{Related Work}
\label{sect:related}

\subsection{Visual Token Compression}

Early studies on visual token compression focus on vision transformers (ViTs), where token pruning~\cite{haurum2023tokens, liang2022evit, kong2022spvit, tang2022ps, yin2022avit, rao2021dynamicvit}, token merging~\cite{bolya2023tome, khaki2024optin}, and compact token representation learning~\cite{wei2023tps} improve throughput by reducing redundant computations. Building on these ideas, many recent methods extend token compression to vision–language models (VLMs). Approaches such as LLaVA-PruMerge~\cite{shang2025prumerge}, HIRED~\cite{arif2025hired}, and VisionZip~\cite{yang2025visionzip} selectively prune redundant visual tokens after the encoder using attention-based salience metrics. However, their query-agnostic nature leads to significant degradation under high sparsity.

To address these limitations, query-aware pruning methods emerge. FastV~\cite{chen2024fastv} uses early LLM attention maps to guide token selection based on the query. SparseVLM~\cite{zhang2025sparsevlm} uses cross-modal attention scores to remove visually irrelevant tokens during prefilling. While these methods preserve accuracy for single-turn queries, they struggle in multi-turn conversations: once a token is pruned, it cannot be recovered for future queries, leading to cumulative information loss. As shown in \fig{fig:query_oracle_degradation}, query-aware pruning exhibits rapid degradation over consecutive rounds, often underperforming query-agnostic methods when the context persists across turns.

This trade-off motivates \method, which decouples sparsity across the inference pipeline. During the prefill stage, \method performs query-agnostic pruning to remove redundant or uninformative visual tokens while retaining a comprehensive visual cache. During decoding, it retrieves only query-relevant tokens from the preserved cache, achieving significant end-to-end acceleration without compromising multi-turn consistency or contextual fidelity. By distributing sparsity between stages, \method combines the generalization of query-agnostic pruning with the adaptivity of query-aware retrieval.

\subsection{KV Cache Compression}

In long-context LLM and VLM inference, the key–value (KV) cache grows linearly with sequence length, imposing substantial latency and memory overhead. Recent work introduces a variety of KV cache compression techniques to address this challenge. StreamingLLM~\cite{xiao2024streamingllm} maintains a finite cache by preserving attention sinks and discarding older context tokens. SnapKV~\cite{li2024snapkv} predicts token importance within an observation window to avoid redundant storage. H2O~\cite{zhang2023h2o} identifies essential KV entries based on cumulative historical attention scores, prioritizing the most influential tokens. Although very effective for text-based applications, these methods often discard cache entries that remain relevant for future decoding steps, which limits their reliability in multi-turn or temporally extended reasoning.

Quest~\cite{tang2024quest} estimates upper-bound attention scores for each page to preserve critical tokens. LazyLLM~\cite{fu2024lazyllm} defers KV computation until the corresponding tokens are required. DuoAttention~\cite{xiao2025duoattention} separates retrieval and streaming heads, assigning full caches to retrieval heads and fixed-length caches to streaming heads. While these strategies efficiently reduce decoding latency for text-only LLMs, they overlook the structured spatial and temporal sparsity inherent to image and video inputs. Visual and video understanding tasks naturally exhibit redundancy across frames and regions, offering further potential for selective retention and retrieval.

\method complements these approaches by integrating visual token sparsity with query-aware KV cache retrieval. Instead of discarding visual context, it preserves a compact but reusable visual cache that supports dynamic retrieval across conversation rounds. This design enables efficient multimodal decoding, sustaining performance under long-context, multi-turn interaction while avoiding the information loss typical of purely text-based compression methods.

\section{Conclusion}
We present \method, a unified sparsity framework that accelerates Vision Language Model (VLM) inference by decoupling visual compression across prefill and decoding. \method prunes redundant visual tokens during prefill and selectively retrieves query-relevant tokens during decoding, reducing latency where it matters most. This design scales from short-context image tasks to long-horizon video and reasoning workloads, where decoding dominates overall inference time. Considering the entire VLM inference stack -- visual embedding, prefill, and decoding -- \method achieves up to \textbf{4.0$\times$} faster prefilling, \textbf{2.5$\times$} faster decoding, and \textbf{2.6$\times$} end-to-end speedup, while preserving or improving accuracy on multi-turn and reasoning benchmarks. Unlike prior pruning methods that trade speed for capability, \method maintains fidelity across modalities and architectures through decoupled sparsity allocation and efficient kernel design. This establishes a scalable, training-free foundation for accelerating the next generation of multimodal systems.

{
\small
\bibliographystyle{unsrt}
\bibliography{main}

\begin{thebibliography}{10}

\bibitem{liu2024llava}
Haotian Liu, Chunyuan Li, Qingyang Wu, and Yong~Jae Lee.
\newblock {Visual Instruction Tuning}.
\newblock In {\em Proceedings of the Conference on Neural Information Processing Systems (NeurIPS)}, 2024.

\bibitem{lin2024vila}
Ji~Lin, Hongxu Yin, Wei Ping, Yao Lu, Pavlo Molchanov, Andrew Tao, Huizi Mao, Jan Kautz, Mohammad Shoeybi, and Song Han.
\newblock {VILA: On Pre-training for Visual Language Models}.
\newblock In {\em Proceedings of the IEEE/CVF Conference on Computer Vision and Pattern Recognition (CVPR)}, 2024.

\bibitem{bai2023qwenvl}
Jinze Bai, Shuai Bai, Shusheng Yang, Shijie Wang, Sinan Tan, Peng Wang, Junyang Lin, Chang Zhou, and Jingren Zhou.
\newblock {Qwen-VL: A Versatile Vision-Language Model for Understanding, Localization, Text Reading, and Beyond}.
\newblock {\em arXiv:2308.12966}, 2023.

\bibitem{wang2024qwen2vl}
Peng Wang, Shuai Bai, Sinan Tan, Shijie Wang, Zhihao Fan, Jinze Bai, Keqin Chen, Xuejing Liu, Jialin Wang, Wenbin Ge, Yang Fan, Kai Dang, Mengfei Du, Xuancheng Ren, Rui Men, Dayiheng Liu, Chang Zhou, Jingren Zhou, and Junyang Lin.
\newblock {Qwen2-VL: Enhancing Vision-Language Model's Perception of the World at Any Resolution}.
\newblock {\em arXiv:2409.12191}, 2024.

\bibitem{liu2025nvila}
Zhijian Liu, Ligeng Zhu, Baifeng Shi, Zhuoyang Zhang, Yuming Lou, Shang Yang, Haocheng Xi, Shiyi Cao, Yuxian Gu, Dacheng Li, Xiuyu Li, Yunhao Fang, Yukang Chen, Cheng-Yu Hsieh, De-An Huang, An-Chieh Cheng, Vishwesh Nath, Jinyi Hu, Sifei Liu, Ranjay Krishna, Daguang Xu, Xiaolong Wang, Pavlo Molchanov, Jan Kautz, Hongxu Yin, Song Han, and Yao Lu.
\newblock {NVILA: Efficient Frontier Visual Language Models}.
\newblock In {\em Proceedings of the IEEE/CVF Conference on Computer Vision and Pattern Recognition (CVPR)}, 2025.

\bibitem{sun2024wanda}
Mingjie Sun, Zhuang Liu, Anna Bair, and J~Zico Kolter.
\newblock {A Simple and Effective Pruning Approach for Large Language Models}.
\newblock In {\em Proceedings of the International Conference on Learning Representations (ICLR)}, 2024.

\bibitem{ma2023llmpruner}
Xinyin Ma, Gongfan Fang, and Xinchao Wang.
\newblock {LLM-Pruner: On the Structural Pruning of Large Language Models}.
\newblock In {\em Proceedings of the Conference on Neural Information Processing Systems (NeurIPS)}, 2023.

\bibitem{zhu2021vtp}
Mingjian Zhu, Yehui Tang, and Kai Han.
\newblock {Vision Transformer Pruning}.
\newblock {\em arXiv:2104.08500}, 2021.

\bibitem{lin2025qserve}
Yujun Lin, Haotian Tang, Shang Yang, Zhekai Zhang, Guangxuan Xiao, Chuang Gan, and Song Han.
\newblock {QServe: W4A8KV4 Quantization and System Co-Design for Efficient LLM Serving}.
\newblock In {\em Proceedings of the Conference on Machine Learning and Systems (MLSys)}, 2025.

\bibitem{hooper2024kvquant}
Coleman Hooper, Sehoon Kim, Hiva Mohammadzadeh, Michael~W Mahoney, Yakun~Sophia Shao, Kurt Keutzer, and Amir Gholami.
\newblock {KVQuant: Towards 10 Million Context Length LLM Inference with KV Cache Quantization}.
\newblock In {\em Proceedings of the Conference on Neural Information Processing Systems (NeurIPS)}, 2024.

\bibitem{liu2025kivi}
Zirui Liu, Jiayi Yuan, Hongye Jin, Shaochen Zhong, Zhaozhuo Xu, Vladimir Braverman, Beidi Chen, and Xia Hu.
\newblock {KIVI: A Tuning-Free Asymmetric 2bit Quantization for KV Cache}.
\newblock In {\em Proceedings of the International Conference on Machine Learning (ICML)}, 2025.

\bibitem{chen2024fastv}
Liang Chen, Haozhe Zhao, Tianyu Liu, Shuai Bai, Junyang Lin, Chang Zhou, and Baobao Chang.
\newblock {An Image is Worth 1/2 Tokens After Layer 2: Plug-and-Play Inference Acceleration for Large Vision-Language Models}.
\newblock In {\em Proceedings of the European Conference on Computer Vision (ECCV)}, 2024.

\bibitem{yang2025visionzip}
Senqiao Yang, Yukang Chen, Zhuotao Tian, Chengyao Wang, Jingyao Li, Bei Yu, and Jiaya Jia.
\newblock {VisionZip: Longer is Better but Not Necessary in Vision Language Models}.
\newblock In {\em Proceedings of the IEEE/CVF Conference on Computer Vision and Pattern Recognition (CVPR)}, 2025.

\bibitem{hudson2019gqa}
Drew~A Hudson and Christopher~D Manning.
\newblock {GQA: A New Dataset for Real-World Visual Reasoning and Compositional Question Answering}.
\newblock In {\em Proceedings of the IEEE/CVF Conference on Computer Vision and Pattern Recognition (CVPR)}, 2019.

\bibitem{bolya2023tome}
Daniel Bolya, Cheng-Yang Fu, Xiaoliang Dai, Peizhao Zhang, Christoph Feichtenhofer, and Judy Hoffman.
\newblock {Token Merging: Your ViT But Faster}.
\newblock In {\em Proceedings of the International Conference on Learning Representations (ICLR)}, 2023.

\bibitem{kong2022spvit}
Zhenglun Kong, Peiyan Dong, Xiaolong Ma, Xin Meng, Mengshu Sun, Wei Niu, Xuan Shen, Geng Yuan, Bin Ren, Minghai Qin, Hao Tang, and Yanzhi Wang.
\newblock {SPViT: Enabling Faster Vision Transformers via Soft Token Pruning}.
\newblock In {\em Proceedings of the European Conference on Computer Vision (ECCV)}, 2022.

\bibitem{khaki2024optin}
Samir Khaki and Konstantinos~N Plataniotis.
\newblock {The Need for Speed: Pruning Transformers with One Recipe}.
\newblock In {\em Proceedings of the International Conference on Learning Representations (ICLR)}, 2024.

\bibitem{chen2023sparsevit}
Xuanyao Chen, Zhijian Liu, Haotian Tang, Li~Yi, Hang Zhao, and Song Han.
\newblock {SparseViT: Revisiting Activation Sparsity for Efficient High-Resolution Vision Transformer}.
\newblock In {\em Proceedings of the IEEE/CVF Conference on Computer Vision and Pattern Recognition (CVPR)}, 2023.

\bibitem{tang2023dtop}
Quan Tang, Bowen Zhang, Jiajun Liu, Fagui Liu, and Yifan Liu.
\newblock {Dynamic Token Pruning in Plain Vision Transformers for Semantic Segmentation}.
\newblock In {\em Proceedings of the IEEE/CVF International Conference on Computer Vision (ICCV)}, 2023.

\bibitem{bolya2023tomesd}
Daniel Bolya and Judy Hoffman.
\newblock {Token Merging for Fast Stable Diffusion}.
\newblock {\em arXiv:2303.17604}, 2023.

\bibitem{kim2022ltp}
Sehoon Kim, Sheng Shen, David Thorsley, Amir Gholami, Woosuk Kwon, Joseph Hassoun, and Kurt Keutzer.
\newblock {Learned Token Pruning for Transformers}.
\newblock In {\em Proceedings of the ACM SIGKDD Conference on Knowledge Discovery and Data Mining (KDD)}, 2022.

\bibitem{zhang2025sparsevlm}
Yuan Zhang, Chun-Kai Fan, Junpeng Ma, Wenzhao Zheng, Tao Huang, Kuan Cheng, Denis Gudovskiy, Tomoyuki Okuno, Yohei Nakata, Kurt Keutzer, and Shanghang Zhang.
\newblock {SparseVLM: Visual Token Sparsification for Efficient Vision-Language Model Inference}.
\newblock In {\em Proceedings of the International Conference on Machine Learning (ICML)}, 2025.

\bibitem{huang2024ivtp}
Kai Huang, Hao Zou, Ye~Xi, BoChen Wang, Zhen Xie, and Liang Yu.
\newblock {IVTP: Instruction-Guided Visual Token Pruning for Large Vision-Language Models}.
\newblock In {\em Proceedings of the European Conference on Computer Vision (ECCV)}, 2024.

\bibitem{shang2025prumerge}
Yuzhang Shang, Mu~Cai, Bingxin Xu, Yong~Jae Lee, and Yan Yan.
\newblock {LLaVA-PruMerge: Adaptive Token Reduction for Efficient Large Multimodal Models}.
\newblock In {\em Proceedings of the IEEE/CVF International Conference on Computer Vision (ICCV)}, 2025.

\bibitem{li2023pope}
Yifan Li, Yifan Du, Kun Zhou, Jinpeng Wang, Wayne~Xin Zhao, and Ji-Rong Wen.
\newblock {Evaluating Object Hallucination in Large Vision-Language Models}.
\newblock In {\em Proceedings of the Conference on Empirical Methods in Natural Language Processing (EMNLP)}, 2023.

\bibitem{arif2025hired}
Kazi Hasan~Ibn Arif, JinYi Yoon, Dimitrios~S Nikolopoulos, Hans Vandierendonck, Deepu John, and Bo~Ji.
\newblock {HiRED: Attention-Guided Token Dropping for Efficient Inference of High-Resolution Vision-Language Models}.
\newblock In {\em Proceedings of the AAAI Conference on Artificial Intelligence (AAAI)}, 2025.

\bibitem{ranzinger2024amradio}
Mike Ranzinger, Greg Heinrich, Jan Kautz, and Pavlo Molchanov.
\newblock {AM-RADIO: Agglomerative Vision Foundation Model -- Reduce All Domains Into One}.
\newblock In {\em Proceedings of the IEEE/CVF Conference on Computer Vision and Pattern Recognition (CVPR)}, 2024.

\bibitem{heinrich2025radio25}
Greg Heinrich, Mike Ranzinger, Hongxu Yin, Yao Lu, Jan Kautz, Andrew Tao, Bryan Catanzaro, and Pavlo Molchanov.
\newblock {RADIOv2.5: Improved Baselines for Agglomerative Vision Foundation Models}.
\newblock In {\em Proceedings of the IEEE/CVF Conference on Computer Vision and Pattern Recognition (CVPR)}, 2025.

\bibitem{tillet2019triton}
Philippe Tillet, Hsiang-Tsung Kung, and David Cox.
\newblock {Triton: An Intermediate Language and Compiler for Tiled Neural Network Computations}.
\newblock In {\em Proceedings of the ACM SIGPLAN International Workshop on Machine Learning and Programming Languages (MAPL)}, 2019.

\bibitem{dao2024flashattention2}
Tri Dao.
\newblock {FlashAttention-2: Faster Attention with Better Parallelism and Work Partitioning}.
\newblock In {\em Proceedings of the International Conference on Learning Representations (ICLR)}, 2024.

\bibitem{sermanet2024robovqa}
Pierre Sermanet, Tianli Ding, Jeffrey Zhao, Fei Xia, Debidatta Dwibedi, Keerthana Gopalakrishnan, Christine Chan, Gabriel Dulac-Arnold, Sharath Maddineni, Nikhil~J Joshi, Pete Florence, Wei Han, Robert Baruch, Yao Lu, Suvir Mirchandani, Peng Xu, Pannag Sanketi, Karol Hausman, Izhak Shafran, Brian Ichter, and Yuan Cao.
\newblock {RoboVQA: Multimodal Long-Horizon Reasoning for Robotics}.
\newblock In {\em Proceedings of the IEEE International Conference on Robotics and Automation (ICRA)}, 2024.

\bibitem{kang2025var}
Seil Kang, Jinyeong Kim, Junhyeok Kim, and Seong~Jae Hwang.
\newblock {See What You Are Told: Visual Attention Sink in Large Multimodal Models}.
\newblock In {\em Proceedings of the International Conference on Learning Representations (ICLR)}, 2025.

\bibitem{xing2025pyramiddrop}
Long Xing, Qidong Huang, Xiaoyi Dong, Jiajie Lu, Pan Zhang, Yuhang Zang, Yuhang Cao, Conghui He, Jiaqi Wang, Feng Wu, and Dahua Lin.
\newblock {PyramidDrop: Accelerating Your Large Vision-Language Models via Pyramid Visual Redundancy Reduction}.
\newblock In {\em Proceedings of the IEEE/CVF Conference on Computer Vision and Pattern Recognition (CVPR)}, 2025.

\bibitem{xiao2023smoothquant}
Guangxuan Xiao, Ji~Lin, Mickael Seznec, Hao Wu, Julien Demouth, and Song Han.
\newblock {SmoothQuant: Accurate and Efficient Post-Training Quantization for Large Language Models}.
\newblock In {\em Proceedings of the International Conference on Machine Learning (ICML)}, 2023.

\bibitem{lin2024awq}
Ji~Lin, Jiaming Tang, Haotian Tang, Shang Yang, Wei-Ming Chen, Wei-Chen Wang, Guangxuan Xiao, Xingyu Dang, Chuang Gan, and Song Han.
\newblock {AWQ: Activation-Aware Weight Quantization for LLM Compression and Acceleration}.
\newblock In {\em Proceedings of the Conference on Machine Learning and Systems (MLSys)}, 2024.

\bibitem{kembhavi2016ai2d}
Aniruddha Kembhavi, Mike Salvato, Eric Kolve, Minjoon Seo, Hannaneh Hajishirzi, and Ali Farhadi.
\newblock {A Diagram Is Worth A Dozen Images}.
\newblock In {\em Proceedings of the European Conference on Computer Vision (ECCV)}, 2016.

\bibitem{masry2022chartqa}
Ahmed Masry, Do~Xuan Long, Jia~Qing Tan, Shafiq Joty, and Enamul Hoque.
\newblock {ChartQA: A Benchmark for Question Answering about Charts with Visual and Logical Reasoning}.
\newblock In {\em Proceedings of the Annual Meeting of the Association for Computational Linguistics (ACL)}, 2022.

\bibitem{mathew2021docvqa}
Minesh Mathew, Dimosthenis Karatzas, and CV~Jawahar.
\newblock {DocVQA: A Dataset for VQA on Document Images}.
\newblock In {\em Proceedings of the IEEE/CVF Winter Conference on Applications of Computer Vision (WACV)}, 2021.

\bibitem{mathew2022infographicvqa}
Minesh Mathew, Viraj Bagal, Rub{\`e}n Tito, Dimosthenis Karatzas, Ernest Valveny, and CV~Jawahar.
\newblock {InfographicVQA}.
\newblock In {\em Proceedings of the IEEE/CVF Winter Conference on Applications of Computer Vision (WACV)}, 2022.

\bibitem{fu2023mme}
Chaoyou Fu, Peixian Chen, Yunhang Shen, Yulei Qin, Mengdan Zhang, Xu~Lin, Jinrui Yang, Xiawu Zheng, Ke~Li, Xing Sun, Yunsheng Wu, and Rongrong Ji.
\newblock {MME: A Comprehensive Evaluation Benchmark for Multimodal Large Language Models}.
\newblock {\em arXiv:2306.13394}, 2023.

\bibitem{lu2022scienceqa}
Pan Lu, Swaroop Mishra, Tanglin Xia, Liang Qiu, Kai-Wei Chang, Song-Chun Zhu, Oyvind Tafjord, Peter Clark, and Ashwin Kalyan.
\newblock {Learn to Explain: Multimodal Reasoning via Thought Chains for Science Question Answering}.
\newblock In {\em Proceedings of the Conference on Neural Information Processing Systems (NeurIPS)}, 2022.

\bibitem{singh2019textvqa}
Amanpreet Singh, Vivek Natarajan, Meet Shah, Yu~Jiang, Xinlei Chen, Dhruv Batra, Devi Parikh, and Marcus Rohrbach.
\newblock {Towards VQA Models That Can Read}.
\newblock In {\em Proceedings of the IEEE/CVF Conference on Computer Vision and Pattern Recognition (CVPR)}, 2019.

\bibitem{liu2024llavanext}
Haotian Liu, Chunyuan Li, Yuheng Li, Bo~Li, Yuanhan Zhang, Sheng Shen, and Yong~Jae Lee.
\newblock {LLaVA-NeXT: Improved Reasoning, OCR, and World Knowledge}.
\newblock \url{https://llava-vl.github.io/blog/2024-01-30-llava-next/}, 2024.

\bibitem{wang2025internvl35}
Weiyun Wang, Zhangwei Gao, Lixin Gu, Hengjun Pu, Long Cui, Xingguang Wei, Zhaoyang Liu, Linglin Jing, Shenglong Ye, Jie Shao, Zhaokai Wang, Zhe Chen, Hongjie Zhang, Ganlin Yang, Haomin Wang, Qi~Wei, Jinhui Yin, Wenhao Li, Erfei Cui, Guanzhou Chen, Zichen Ding, Changyao Tian, Zhenyu Wu, Jingjing Xie, Zehao Li, Bowen Yang, Yuchen Duan, Xuehui Wang, Zhi Hou, Haoran Hao, Tianyi Zhang, Songze Li, Xiangyu Zhao, Haodong Duan, Nianchen Deng, Bin Fu, Yinan He, Yi~Wang, Conghui He, Botian Shi, Junjun He, Yingtong Xiong, Han Lv, Lijun Wu, Wenqi Shao, Kaipeng Zhang, Huipeng Deng, Biqing Qi, Jiaye Ge, Qipeng Guo, Wenwei Zhang, Songyang Zhang, Maosong Cao, Junyao Lin, Kexian Tang, Jianfei Gao, Haian Huang, Yuzhe Gu, Chengqi Lyu, Huanze Tang, Rui Wang, Haijun Lv, Wanli Ouyang, Limin Wang, Min Dou, Xizhou Zhu, Tong Lu, Dahua Lin, Jifeng Dai, Weijie Su, Bowen Zhou, Kai Chen, Yu~Qiao, Wenhai Wang, and Gen Luo.
\newblock {InternVL3.5: Advancing Open-Source Multimodal Models in Versatility, Reasoning, and Efficiency}.
\newblock {\em arXiv:2508.18265}, 2025.

\bibitem{nvidia2025nemotronnanovl}
NVIDIA.
\newblock {Llama Nemotron Nano VL}.
\newblock \url{https://build.nvidia.com/nvidia/llama-3.1-nemotron-nano-vl-8b-v1}, 2025.

\bibitem{wu2024longvideobench}
Haoning Wu, Dongxu Li, Bei Chen, and Junnan Li.
\newblock {LongVideoBench: A Benchmark for Long-Context Interleaved Video-Language Understanding}.
\newblock In {\em Proceedings of the Conference on Neural Information Processing Systems (NeurIPS)}, 2024.

\bibitem{zhou2025mlvu}
Junjie Zhou, Yan Shu, Bo~Zhao, Boya Wu, Zhengyang Liang, Shitao Xiao, Minghao Qin, Xi~Yang, Yongping Xiong, Bo~Zhang, Tiejun Huang, and Zheng Liu.
\newblock {MLVU: Benchmarking Multi-Task Long Video Understanding}.
\newblock In {\em Proceedings of the IEEE/CVF Conference on Computer Vision and Pattern Recognition (CVPR)}, 2025.

\bibitem{xiao2021nextqa}
Junbin Xiao, Xindi Shang, Angela Yao, and Tat-Seng Chua.
\newblock {NExT-QA: Next Phase of Question-Answering to Explaining Temporal Actions}.
\newblock In {\em Proceedings of the IEEE/CVF Conference on Computer Vision and Pattern Recognition (CVPR)}, 2021.

\bibitem{fu2025videomme}
Chaoyou Fu, Yuhan Dai, Yongdong Luo, Lei Li, Shuhuai Ren, Renrui Zhang, Zihan Wang, Chenyu Zhou, Yunhang Shen, Mengdan Zhang, Peixian Chen, Yanwei Li, Shaohui Lin, Sirui Zhao, Ke~Li, Tong Xu, Xiawu Zheng, Enhong Chen, Caifeng Shan, Ran He, and Xing Sun.
\newblock {Video-MME: The First-Ever Comprehensive Evaluation Benchmark of Multi-Modal LLMs in Video Analysis}.
\newblock In {\em Proceedings of the IEEE/CVF Conference on Computer Vision and Pattern Recognition (CVPR)}, 2025.

\bibitem{chen2025longvila}
Yukang Chen, Fuzhao Xue, Dacheng Li, Qinghao Hu, Ligeng Zhu, Xiuyu Li, Yunhao Fang, Haotian Tang, Shang Yang, Zhijian Liu, Ethan He, Hongxu Yin, Pavlo Molchanov, Jan Kautz, Linxi Fan, Yuke Zhu, Yao Lu, and Song Han.
\newblock {LongVILA: Scaling Long-Context Visual Language Models for Long Videos}.
\newblock In {\em Proceedings of the International Conference on Learning Representations (ICLR)}, 2025.

\bibitem{bai2025qwen25vl}
Shuai Bai, Keqin Chen, Xuejing Liu, Jialin Wang, Wenbin Ge, Sibo Song, Kai Dang, Peng Wang, Shijie Wang, Jun Tang, Humen Zhong, Yuanzhi Zhu, Mingkun Yang, Zhaohai Li, Jianqiang Wan, Pengfei Wang, Wei Ding, Zheren Fu, Yiheng Xu, Jiabo Ye, Xi~Zhang, Tianbao Xie, Zesen Cheng, Hang Zhang, Zhibo Yang, Haiyang Xu, and Junyang Lin.
\newblock {Qwen2.5-VL Technical Report}.
\newblock {\em arXiv:2502.13923}, 2025.

\bibitem{xiao2024streamingllm}
Guangxuan Xiao, Yuandong Tian, Beidi Chen, Song Han, and Mike Lewis.
\newblock {Efficient Streaming Language Models with Attention Sinks}.
\newblock In {\em Proceedings of the International Conference on Learning Representations (ICLR)}, 2024.

\bibitem{maaz2023videochatgpt}
Muhammad Maaz, Hanoona Rasheed, Salman Khan, and Fahad~Shahbaz Khan.
\newblock {Video-ChatGPT: Towards Detailed Video Understanding via Large Vision and Language Models}.
\newblock In {\em Proceedings of the Annual Meeting of the Association for Computational Linguistics (ACL)}, 2023.

\bibitem{zhang2024longva}
Peiyuan Zhang, Kaichen Zhang, Bo~Li, Guangtao Zeng, Jingkang Yang, Yuanhan Zhang, Ziyue Wang, Haoran Tan, Chunyuan Li, and Ziwei Liu.
\newblock {Long Context Transfer from Language to Vision}.
\newblock {\em arXiv:2406.16852}, 2024.

\bibitem{chen2025longvilar1}
Yukang Chen, Wei Huang, Baifeng Shi, Qinghao Hu, Hanrong Ye, Ligeng Zhu, Zhijian Liu, Pavlo Molchanov, Jan Kautz, Xiaojuan Qi, Sifei Liu, Hongxu Yin, Yao Lu, and Song Han.
\newblock {Scaling RL to Long Videos}.
\newblock {\em arXiv:2507.07966}, 2025.

\bibitem{yang2024qwen2}
An~Yang, Baosong Yang, Binyuan Hui, Bo~Zheng, Bowen Yu, Chang Zhou, Chengpeng Li, Chengyuan Li, Dayiheng Liu, Fei Huang, Guanting Dong, Haoran Wei, Huan Lin, Jialong Tang, Jialin Wang, Jian Yang, Jianhong Tu, Jianwei Zhang, Jianxin Ma, Jianxin Yang, Jin Xu, Jingren Zhou, Jinze Bai, Jinzheng He, Junyang Lin, Kai Dang, Keming Lu, Keqin Chen, Kexin Yang, Mei Li, Mingfeng Xue, Na~Ni, Pei Zhang, Peng Wang, Ru~Peng, Rui Men, Ruize Gao, Runji Lin, Shijie Wang, Shuai Bai, Sinan Tan, Tianhang Zhu, Tianhao Li, Tianyu Liu, Wenbin Ge, Xiaodong Deng, Xiaohuan Zhou, Xingzhang Ren, Xinyu Zhang, Xipin Wei, Xuancheng Ren, Xuejing Liu, Yang Fan, Yang Yao, Yichang Zhang, Yu~Wan, Yunfei Chu, Yuqiong Liu, Zeyu Cui, Zhenru Zhang, Zhifang Guo, and Zhihao Fan.
\newblock {Qwen2 Technical Report}.
\newblock {\em arXiv:2407.10671}, 2024.

\bibitem{haurum2023tokens}
Joakim~Bruslund Haurum, Sergio Escalera, Graham~W Taylor, and Thomas~B Moeslund.
\newblock {Which Tokens to Use? Investigating Token Reduction in Vision Transformers}.
\newblock In {\em Proceedings of the IEEE/CVF International Conference on Computer Vision (ICCV)}, 2023.

\bibitem{liang2022evit}
Youwei Liang, Chongjian Ge, Zhan Tong, Yibing Song, Jue Wang, and Pengtao Xie.
\newblock {Not All Patches are What You Need: Expediting Vision Transformers via Token Reorganizations}.
\newblock In {\em Proceedings of the International Conference on Learning Representations (ICLR)}, 2022.

\bibitem{tang2022ps}
Yehui Tang, Kai Han, Yunhe Wang, Chang Xu, Jianyuan Guo, Chao Xu, and Dacheng Tao.
\newblock {Patch Slimming for Efficient Vision Transformers}.
\newblock In {\em Proceedings of the IEEE/CVF Conference on Computer Vision and Pattern Recognition (CVPR)}, 2022.

\bibitem{yin2022avit}
Hongxu Yin, Arash Vahdat, Jose~M Alvarez, Arun Mallya, Jan Kautz, and Pavlo Molchanov.
\newblock {A-ViT: Adaptive Tokens for Efficient Vision Transformer}.
\newblock In {\em Proceedings of the IEEE/CVF Conference on Computer Vision and Pattern Recognition (CVPR)}, 2022.

\bibitem{rao2021dynamicvit}
Yongming Rao, Wenliang Zhao, Benlin Liu, Jiwen Lu, Jie Zhou, and Cho-Jui Hsieh.
\newblock {DynamicViT: Efficient Vision Transformers with Dynamic Token Sparsification}.
\newblock In {\em Proceedings of the Conference on Neural Information Processing Systems (NeurIPS)}, 2021.

\bibitem{wei2023tps}
Siyuan Wei, Tianzhu Ye, Shen Zhang, Yao Tang, and Jiajun Liang.
\newblock {Joint Token Pruning and Squeezing Towards More Aggressive Compression of Vision Transformers}.
\newblock In {\em Proceedings of the IEEE/CVF Conference on Computer Vision and Pattern Recognition (CVPR)}, 2023.

\bibitem{li2024snapkv}
Yuhong Li, Yingbing Huang, Bowen Yang, Bharat Venkitesh, Acyr Locatelli, Hanchen Ye, Tianle Cai, Patrick Lewis, and Deming Chen.
\newblock {SnapKV: LLM Knows What You are Looking for Before Generation}.
\newblock In {\em Proceedings of the Conference on Neural Information Processing Systems (NeurIPS)}, 2024.

\bibitem{zhang2023h2o}
Zhenyu Zhang, Ying Sheng, Tianyi Zhou, Tianlong Chen, Lianmin Zheng, Ruisi Cai, Zhao Song, Yuandong Tian, Christopher R{\'e}, Clark Barrett, Zhangyang Wang, and Beidi Chen.
\newblock {H2O: Heavy-Hitter Oracle for Efficient Generative Inference of Large Language Models}.
\newblock In {\em Proceedings of the Conference on Neural Information Processing Systems (NeurIPS)}, 2023.

\bibitem{tang2024quest}
Jiaming Tang, Yilong Zhao, Kan Zhu, Guangxuan Xiao, Baris Kasikci, and Song Han.
\newblock {Quest: Query-Aware Sparsity for Efficient Long-Context LLM Inference}.
\newblock In {\em Proceedings of the International Conference on Machine Learning (ICML)}, 2024.

\bibitem{fu2024lazyllm}
Qichen Fu, Minsik Cho, Thomas Merth, Sachin Mehta, Mohammad Rastegari, and Mahyar Najibi.
\newblock {LazyLLM: Dynamic Token Pruning for Efficient Long Context LLM Inference}.
\newblock {\em arXiv:2407.14057}, 2024.

\bibitem{xiao2025duoattention}
Guangxuan Xiao, Jiaming Tang, Jingwei Zuo, Junxian Guo, Shang Yang, Haotian Tang, Yao Fu, and Song Han.
\newblock {DuoAttention: Efficient Long-Context LLM Inference with Retrieval and Streaming Heads}.
\newblock In {\em Proceedings of the International Conference on Learning Representations (ICLR)}, 2025.

\end{thebibliography}
}

\end{document}